\pdfoutput=1

\documentclass[11pt]{article}
\usepackage{ACL2023}
\usepackage{times}
\usepackage{latexsym}
\usepackage{graphicx}
\usepackage{longtable}
\usepackage{makecell} 

\usepackage[T1]{fontenc}

\usepackage[utf8]{inputenc}
\usepackage{amsmath}
\usepackage{cleveref}

\usepackage{microtype}

\usepackage{inconsolata}
\usepackage{multirow}

\usepackage{booktabs}
\usepackage{makecell}
\usepackage{tabularx}
\usepackage{soul}

\usepackage{newunicodechar}
\newunicodechar{ً}{}

\definecolor{goalgreen}{RGB}{153,204,0}
\definecolor{customblue}{RGB}{0,0,153}

\usepackage[suppress]{color-edits}
\addauthor{ZT}{olive}

\usepackage{enumerate}
\usepackage[shortlabels]{enumitem}
\usepackage{amssymb}

\usepackage[]{todonotes}
\makeatletter
\newcommand*\iftodonotes{\if@todonotes@disabled\expandafter\@secondoftwo\else\expandafter\@firstoftwo\fi}  
\newcommand{\noindentaftertodo}{\iftodonotes{\noindent}{}}

\newcommand{\note}[4][]{\todo[author=#2,color=#3,size=\scriptsize,caption={},#1]{#4}} 
\everypar{\looseness=-1}

\newcommand{\zc}[2][]{\note[#1]{ZT}{olive!40}{#2}}
\newcommand{\ZC}[2][]{\zc[inline,#1]
{#2}\noindentaftertodo}

\usepackage[table]{xcolor}

\definecolor{seamecolor}{HTML}{a93226}
\definecolor{proprietarydatacolor}{HTML}{000000}
\definecolor{asru2019color}{HTML}{e74c3c}
\definecolor{talcscorpuscolor}{HTML}{E97451}
\definecolor{ascendcolor}{HTML}{F88379}
\definecolor{magicdataramccolor}{HTML}{FAA0A0}
\definecolor{asru2021color}{HTML}{E0115F}
\definecolor{aicubescompetitiondatasetcolor}{HTML}{f39c12}
\definecolor{iscslp2022magichubcodeswitchingasrchallengedatasetcolor}{HTML}{b9770e}
\definecolor{ntutab01color}{HTML}{fdedec}

\definecolor{iiithhecmcolor}{HTML}{98FB98}
\definecolor{iitghingcoscolor}{HTML}{B4C424}
\definecolor{mucscolor}{HTML}{00A36C}
\definecolor{vitbhebiccolor}{HTML}{8fbc8f}

\definecolor{escwacolor}{HTML}{00fa9a}
\definecolor{arzncolor}{HTML}{3498db}
\definecolor{qasr.cscolor}{HTML}{7fb3d5}
\definecolor{haccolor}{HTML}{9467BD}
\definecolor{tunswitchcolor}{HTML}{40E0D0}
\definecolor{mixatcolor}{HTML}{0000FF}
\definecolor{zaebucspokencolor}{HTML}{d6eaf8}
\definecolor{casablancacolor}{HTML}{8C564B}
\definecolor{saudilangcorpus(scc)(evalonly)color}{HTML}{808080}

\definecolor{germanspokenwikipediacorpuscolor}{HTML}{CF9FFF}
\definecolor{decmcolor}{HTML}{E377C2}
\definecolor{medreccolor}{HTML}{FFBF00}
\definecolor{kecscolor}{HTML}{FFEA00}
\definecolor{bangormiamicolor}{HTML}{ECFFDC}
\definecolor{lotusbicolor}{HTML}{7F7F7F}
\definecolor{hartantocscolor}{HTML}{1e90ff}
\definecolor{southafricansoapoperascolor}{HTML}{ffff54}
\definecolor{twosepedienglishcsspeechcorpus}{HTML}{7fff00}

\newcommand{\dsbox}[1]{\textcolor{#1}{\rule{1.5ex}{1.5ex}}}
\newcommand{\dsentry}[2]{(\dsbox{#1}, #2)}

\title{Code-Switching in End-to-End Automatic Speech Recognition: \\ A Systematic Literature Review}

\author{Maha Tufail Agro$^{1}$ \quad Atharva Kulkarni$^1$ \quad Karima Kadaoui$^1$ \\  \bf Zeerak Talat$^2$ \quad Hanan Aldarmaki$^1$ \\ $^1$Mohamed bin Zayed University of Artificial Intelligence, UAE \\
    $^2$University of Edinburgh, UK\\
     $^1$\texttt{\{maha.agro, hanan.aldarmaki\}@mbzuai.ac.ae}, $^2$\texttt{z@zeerak.org}}

\date{}

\begin{document}
\maketitle
\begin{abstract}
Motivated by a growing research interest into automatic speech recognition (ASR), and the growing body of work for languages in which code-switching (CS) often occurs, we present a systematic literature review of code-switching in end-to-end ASR models.
We collect and manually annotate papers published in peer reviewed venues. 
We document the languages considered, datasets, metrics, model choices, and performance, 
and present a discussion of challenges in end-to-end ASR for code-switching. 
Our analysis thus provides insights on current research efforts and available resources as well as opportunities and gaps to guide future research.
\end{abstract}

\section{Introduction}
\ZTedit{Automatic speech recognition (ASR) is one of the most widely used technologies for accessible communication and device interaction. 
A recent market analysis estimated the global market of speech and voice technologies at USD 20 billion in 2023, of which ASR accounts for over 60\%, distributed across a range of sectors, including automotive, banking and financial, healthcare, and retail~\cite{grandview2024voice}. 
Such widespread adoption of ASR highlights the need and desire for high performing systems across different languages and speech patterns. 
While large-scale resources for developing robust ASR systems are common for monolingual settings, datasets and methods for code-switching (CS)---the practice of alternating between two or more languages in a single conversation or discourse---remains a challenge for current ASR models, despite that it is a widespread phenomenon across the globe~\cite{UNESCO_Multilingual_2024}. 
Multilingual models may be well suited for handling CS~\cite{peng2023prompting,polywer}, but their performance across language pairs has yet to be fully explored. 
Studies that seek to address CS in speech are limited by a scarcity of datasets that are designed to evaluate this phenomenon. 
In this paper, we present a comprehensive review of efforts towards end-to-end (E2E) ASR systems for CS. 
We seek to identify the current trends in research by answering the following research questions: (1) Which language pairs, modeling, and evaluation mechanisms have been pursued? (2) What is the current state-of-the-art for E2E ASR for CS? (3) What are the persisting challenges in ASR for CS research?
We find that although the body of work in E2E ASR for code-switching is growing, a small subset of languages receive the most attention, and while the number of datasets is growing, there is a large disparity in the languages that are covered by the datasets. 
Moreover, we find a wide variety of training and evaluation methodologies, but the efforts are mostly sporadic, with no consistent benchmarking or clarity on directions for future research. 
To this end, we present a discussion of these challenges around three axes: resources, methodology, and fairness.
Through this work, we hope to provide a road map for more inclusive, rigorous, and reproducible research in end-to-end ASR for code-switching.}

\ZTdelete{This widespread adoption of the technology in critical areas illustrates its practical uses and calls for diversifying the coverage of these models to more spoken varieties. 
Despite the accelerated advances in ASR technology in recent years, code-switching remains a challenging problem for current ASR models. 
Code-switching (CS) refers to the phenomenon of alternating between two or more languages or dialects in a single conversation or discourse. 
It is a natural and widespread feature of communication among multilingual speakers, who make up more than half of the world’s population\footnote{\href{https://www.unesco.org/en/articles/multilingual-education-key-quality-and-inclusive-learning}{UNESCO: Multilingual Education}; last accessed in May 30, 2025.}. 
Yet, most large-scale datasets for training robust ASR models are monolingual. 
Multilingual models may be well suited for handling code-switching \cite{peng2023prompting, polywer} but the extent of their performance across language pairs hasn't been fully explored. 
Studies that address code-switching phenomena in speech are limited by the scarcity of datasets that are specifically designed to evaluate CS utterances, but there has been a recent surge of interest with many new datasets and methodologies proposed to improve the recognition of CS speech. 
In this article, we provide a comprehensive review of recent research efforts into CS in end-to-end ASR systems, providing a reference and road map for future research. } 
\subsection{Code-switching}
\ZTedit{People engage in CS in their communication for many and varied reasons, such as a speaker's familiarity with the setting and language and how they wish to project themselves~\cite{myers-scotton_social_1995}; to express identity or tone~\cite{Gumperz_1982}; and due to diglossia, the presence of two languages or varieties used under different conditions, such as `high' and `low' (colloquial) varieties \cite{Ferguson01011959}. 
CS can also be a consequence of ASR systems themselves; a recent study found that older African-Americans engaged in a form of code-switching while addressing ASR systems that aren't trained to process their vernacular~\cite{Harrington_Its_2022}. In code-switching research, there is a distinction made between the \textit{matrix} language---the language which provides the structure---and the \textit{embedded} language---which provides the foreign words or phrases~\cite{Weller2022}.}
Code-switching is often categorized into two types: \textit{inter-sentential} (switching languages or dialects between sentences) and \textit{intra-sentential} code-switching, where language switches happen within a single sentence. 
The latter is most challenging for speech systems and what is typically addressed in the surveyed literature. 
\ZTdelete{
The main language providing the structure is called the \textit{matrix language}, while the language providing foreign words or phrases, maintaining their original form, is called the \textit{embedded language} \cite{Weller2022}.}

\subsection{Automatic Speech Recognition}
ASR research and development has a long history dating back to the 1950s. \citet{AnusuyaKatti2009} summarize the progress made until 2009; by then, the SOTA models were hybrid HMM-DNN models, combining the controlled statistical power of HMMs for lexical and language modeling with the adaptive nature of neural networks for acoustic modeling. Subsequent research led to the development of fully connected \textit{end-to-end} (or E2E for short) ASR systems. As pointed out in \citet{prabhavalkar2023end}, the term E2E hides various practical complexities in these systems, and it is an umbrella term used to describe various neural ASR systems that are characterized (with caveats) as \textit{joint modeling and training of ASR components from scratch in a single computational graph and objective function}. Refer to \citet{prabhavalkar2023end} for a comprehensive review of these systems and their types. Relevant to our discussion here is that these systems are currently the state-of-the-art in speech recognition, enhanced with self-supervised learning (SSL) methods that utilize large unlabeled data for better generalization. A review of these SSL methods for speech is provided in \citet{mohamed2022self}. These systems achieve remarkable recognition performance across speakers, noise conditions, and even across different languages \cite{yadav-sitaram-2022-survey}.

\subsection{Related Work}

\ZTedit{While there are hundreds of  reviews of ``speech recognition'' systems~\cite[e.g.,][]{reddy1976speech, prabhavalkar2023end}, we identify only one published survey on code-switching in ASR systems~\cite{mustafa2022code}. 
In their study, \citet{mustafa2022code} analyze a sample of papers on bilingual and multilingual ASR, including 24 papers on CS. Their search methodology differs from ours in both focus and scope, as they analyze a relatively small subset of CS papers, many of which predate the ones in our analysis.  
Meanwhile, in text processing \citet{winata-etal-2023-decades} present a comprehensive survey of code-switching, covering decades of research in NLP. 
Our work complements these efforts by focusing exclusively on E2E ASR, which currently represents the mainstream in ASR technology. 
Furthermore, our analysis includes all papers that fit our search and inclusion criteria, which is constrained only by the recency of the studied E2E systems; indeed, over half of the papers in our survey have been published since 2022. 
Our survey thus provides a comprehensive and up-to-date overview of the research on CS in E2E ASR.}
\ZTdelete{A search of review articles where ``speech recognition" appears in the title returns over 1000 articles covering various aspects of speech recognition across the years. 
Searching for terms related to code-switching returns only one published survey \cite{mustafa2022code} and one preprint \cite{sitaram2019survey}. 
In text-processing, \citet{winata-etal-2023-decades} provides a comprehensive survey of code-switching, covering decades of research in NLP. 
Our survey complements these efforts and differs from them by
focusing exclusively on end-to-end (E2E) ASR.  This type of ASR system, which is currently mainstream, has gained popularity relatively recently;  over half of the covered articles in our survey were published in 2022 or later.  
Thus, the survey provides a comprehensive and up-to-date picture of code-switching in E2E ASR, which we hope to illuminate fruitful pathways for future research.}

\subsection{Outline}
We present a systematic literature review of published research on E2E ASR for code-switched speech.
We analyze the coverage of languages, datasets, metrics, and modeling choices, thereby presenting a comprehensive overview of the field. 
We first describe the scope of the study and our search and annotation methodologies in \Cref{sec:sample}, and describe the results of our annotation in the subsequent sections. 
\Cref{section:langs_data} summarizes the languages and datasets covered in this literature.  
We then detail ASR modeling choices in Section \ref{section:asr_modeling}, including architectures, language identification, and decoding strategies, among other dimensions identified in our annotation scheme. Section \ref{section:training} describes the training and evaluation settings used in these papers, such as data augmentation and evaluation metrics. 
We summarize the state-of-the-art models in Section \ref{sec:sota}, and describe new and persisting challenges in \Cref{section:challenges}.

\ZTdelete{We include all relevant and peer-reviewed papers in our analysis, resulting in a comprehensive picture of the state of research in this area. 
In the following sections, we describe our survey scope, methodology, and results. 
Section \ref{sec:data} describes the datasets and languages covered in the literature, including shared tasks.  
Section \ref{sec:methods} summarizes the modeling choices in these papers, including architectures, data augmentation strategies, use of language IDs, choice of text units, use of pretrained models, language models, decoding strategies, and other topics. 
Section \ref{sec:metrics} describes the metrics used to summarize ASR performance in the presence of code-switching, and finally, section \ref{sec:results} summarizes the results and modeling choices of the state-of-the-art models across datasets.}

\begin{table*}[]
\centering
\resizebox{0.8\textwidth}{!}{
\begin{tabular}{@{}lll@{}}
\toprule
\textbf{Dimension}  & \textbf{Attribute} & \textbf{Description} \\ \midrule
\multirow{3}{*}{Problem Setup \& Data} 
 & Language(s) & \textit{Languages studied (see full list in Appendix)} \\
 & Datasets & \textit{Datasets used (see full list in Appendix)} \\
 & Dataset Accessibility & Yes | No \\
 \midrule
\multirow{9}{*}{Model Design Choices} 
 & Monolingual Modeling & Yes | No \\
 & Multilingual Modeling & Yes | No \\
 & LID & Yes | No \\
 & Text Units & Words | Subwords (BPE) | Characters | Phones \\
 & Architecture & \textit{Type of neural architecture used for ASR} \\
 & Pretrained models & Yes | No \\
 & {Language Model} & Yes | No \\
 & Loss Function & CTC | Cross Entropy | Hybrid CTC/Attention | Others \\
 & Decoding Strategy & Greedy | Beam search | Other \\ \midrule
\multirow{4}{*}{Training and Evaluation Settings} 
 & Data augmentation & \textit{Type of data augmentation used} \\
 & Translation & Yes | No \\
 & Zero-Shot & Yes | No \\
 & Evaluation metrics & WER | CER | MER | TER | \textit{etc. }\\ \midrule 
\multirow{2}{*}{Performance} 
 & Best performance & \textit{Best reported performance} \\
 & Best model & \textit{Model with best reported performance} \\ \bottomrule
\end{tabular}
}
\caption{List of annotated attributes in the survey.}
\label{tab:annotation_attributes}
\end{table*}

\section{Scope \& Methodology}
\label{sec:sample}

\paragraph{Data Collection} 
We gathered related papers by querying the Semantic Scholar API\footnote{\url{https://www.semanticscholar.org/product/api}} to retrieve papers that use terms related to code-switching, published from 2014 until February 27, 2025\ZTdelete{(the date of the last query)}.\footnote{We used the following query for retrieving papers: (ASR | Speech recognition) \& (code-switch* | codeswitch* | code switch* | code-mix* | codemix*).}
We initially retrieved 378 papers, from which we only kept papers that have been published in peer-reviewed venues and describe end-to-end code-switching systems, resulting in a set of 127 papers published between 2018 and 2024 for analysis.\footnote{Based on our search results, the first paper related to ASR for CS using an E2E architecture was published in 2018.} 

\paragraph{Annotation} 
The papers were annotated by five annotators after agreeing on the annotation dimensions. 
\ZTedit{The annotations were carried out during scheduled annotation sessions, where any misunderstandings and questions were addressed during the session to ensure consistency in annotation procedures.}
\ZTdelete{The annotations were carried out over several Zoom meetings to resolve any confusion or misunderstanding over annotation types.} 
\ZTedit{Each paper was annotated by one annotator by manually going over the paper text.}
\ZTdelete{Each paper was annotated individually by manually going over the paper text.} 
\ZTedit{We extracted information from each paper across four major dimensions:}
\ZTdelete{We annotated each paper across four major dimensions (see \Cref{tab:annotation_attributes}):}

\begin{itemize}
    \item \textbf{Problem Setup \& Data}:
    We recorded the languages covered (e.g., Mandarin-English, Spanish-English), the datasets used (e.g., SEAME, Bangor Miami corpus) 
    for training and evaluation, and dataset accessibility.\footnote{We understand an `accessible' dataset to be one where a link is made available, it can be obtained upon request, or it is used in subsequent research.}

    \item \textbf{Model Design Choices}:
     We annotated whether monolingual or multilingual components are explicitly modeled, and note whether code-switched data are used for training. 
    \ZTdelete{We annotated the use of monolingual resources (e.g., pretraining on English LibriSpeech), multilingual resources (e.g., pretraining on multilingual CommonVoice), or code-switched-only training data.
    We also recorded whether language identification (LID) was incorporated either as an auxiliary task or during decoding.  We annotated whether language identification (LID) was incorporated either as an auxiliary task or during decoding. }
    We also annotated the text units used (e.g., characters, phones, BPE), the model architecture (e.g., hybrid CTC/attention-based model with Transformer encoder), use of large pretrained models (e.g., Whisper), external language model integration (e.g., shallow fusion during decoding), loss function and the decoding strategies employed (e.g., beam search).
    
    \item \textbf{Training and Evaluation Settings}:
    We annotated whether the study applied any form of data augmentation (e.g., speed perturbation, synthetic code-switching, etc.), translation (e.g., translating code-switched utterances into monolingual text), and zero-shot evaluation (i.e., evaluating multilingual models on code-switched utterances without fine-tuning). 
    We also annotated the evaluation metrics reported (e.g., WER, CER).

    \item \textbf{Performance}: 
    We annotated the best reported result in each paper and the model that achieved that best performance. 
\end{itemize}

\begin{table}[ht]
    \centering
    \scalebox{0.8}{
    \begin{tabular}{lc}
         \toprule
         \textbf{Type} & \textbf{Count} \\
         \midrule
         Modeling & 88 \\
         New Dataset & 19 \\
         Data Augmentation & 10 \\
         New Evaluation Metric & 3 \\
         Shared Task & 2 \\
         Other & 6 \\
         \bottomrule
    \end{tabular}}
    \caption{Number of papers by category.}
    \label{tab:contribution_type}
\end{table}

\ZTdelete{See Table \ref{tab:annotation_attributes} for the full list of attributes.} 
We also coded the type of contributions in each paper \ZTedit{(see \Cref{tab:contribution_type}).} The majority of the surveyed papers describe empirical research focused on model design choices, such as architecture or training methodology. 
The `Other' category includes papers that do not fit the other categories, such as papers that examine text encodings, data partitions, or analysis of existing models.

\section{Languages \& Datasets} \label{section:langs_data}
\begin{table*}[!ht]
    \centering
    \scalebox{0.7}{
    \begin{tabular}{lcccccc}
        \toprule
        \multirow{3}{*}{\textbf{Dataset}} & \multirow{3}{*}{\textbf{Matrix Language}} & \multirow{3}{*}{\textbf{Speech Type}} & \multirow{2}{*}{\textbf{Total}} & \multicolumn{3}{c}{\textbf{Code-Switched Speech}}\\
        & & & \multirow{2}{*}{\textbf{ (hrs)}} & \multicolumn{3}{c}{\textbf{Duration (hr)}} \\
        \cline{5-7}
         &  &  & & \textbf{Train} & \textbf{Dev} & \textbf{Test} \\
         \midrule
        TALCS \cite{li22j_interspeech} & zho & Spontaneous & 587.0 & 555.9 & 8.0 & 23.6 \\
        ASCEND \cite{lovenia2021ascend} & zho & Spontaneous & 10.6 & 8.8 & 0.9 & 0.9 \\
        ASRU 2019 \cite{shi2020asru} & zho & Read & 740.0 & 200.0 & 40.0 & - \\
        SEAME \cite{lyu2010seame} & zho & Spontaneous & 192.0 & 101.1 & 11.4 & - \\
        KSC2 \cite{mussakhojayeva22_interspeech} & kaz & Read & 1127.9 & 26.3 & 0.5 & 0.5 \\
        Hartanto CS \cite{roosadi_handling_2023} & ind & - & 7.15 & 6.0 & - & 1.15 \\
        VITB-HEBiC \cite{jain2024vitb} & hin & Read & 7.5 & - & - & - \\
        MUCS 2021 \cite{diwan21_interspeech} & hin, ben & Spontaneous & 600.0 & 135.9 & & 12.2 \\
        IITG-HingCos \cite{ganji2019iitg} & hin & Read & 25.0 & - & - & - \\
        Mixat \cite{ali2024mixat} & ara & Spontaneous & 15.0 & - & - & - \\
        ZAEBUC-Spoken \cite{hamed2024zaebuc} & ara & Spontaneous & 12.0 & - & - & - \\
        TunSwitch CS \cite{abdallah2024leveraging} & ara, fra & Spontaneous & 163.62 & 8.5 & 0.15 & 0.25 \\
        HAC \cite{hamed2023benchmarking} & ara & Spontaneous & 2.0 & - & - & - \\
        ESCWA-CS \cite{chowdhury2021towards} & ara & Spontaneous & 2.8 & - & - & - \\
        QASR.CS \cite{mubarak2021qasr} & ara & Read & 5.9 & - & - & - \\
        ArzEn \cite{hamed2020arzen} & ara & Spontaneous & 12.0 & -  & -  & -  \\
        FAME \cite{fame} & fry-nld & Spontaneous & 14 & 11.5 & 1.2 & 1.2 \\
        DECM \cite{ugan-etal-2024-decm} & deu & Spontaneous  & 3.38 & - & - & 3.38 \\
        German Spoken Wikipedia Corpus \cite{khosravani_evaluation_2021} & deu & Read & 34 & - & - & - \\
        Miami Bangor \cite{DeucharDaviesHerringCoutoCarter+2014+93+110} & spa & Spontaneous & 35.66 & - & - & - \\
        South African Soap Operas \cite{van_der_westhuizen_first_2018} & xho, sot, tsn, zul & Spontaneous & 14.3 & 2.5 & 0.2 & 0.7 \\
        Two Sepedi SPCS Corpus\cite{modipa2022two} & nso & Read & 10 & -& -\\
        \bottomrule
    \end{tabular}
    }
    \caption{Comprehensive list of publicly available datasets found in the surveyed literature. Embedded language is English unless otherwise specified.}
    \label{tab:dataset_duration_complete}
\end{table*}

\begin{figure} [http!]
\vspace{-1em}
    \centering
    \makebox[0.5\textwidth][c]{\hspace{-0.2cm}\includegraphics[width=0.55\textwidth]{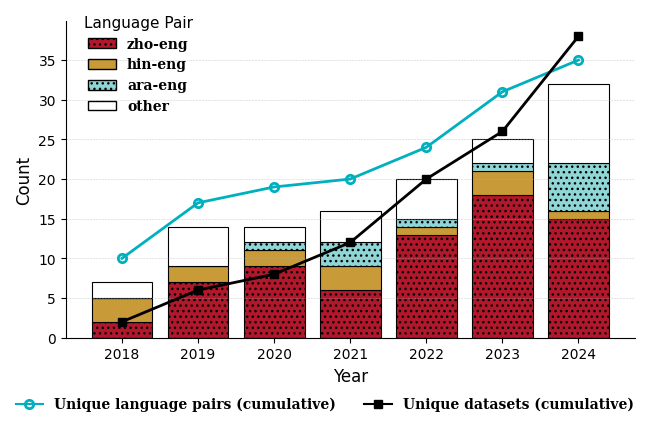}}
   \caption{Total number of papers per year, and the number of papers on each of the top 3 language-pairs. The lines show the cumulative total of unique language pairs and unique datasets covered over time.
   }
    \label{fig:lang_dist}
\vspace{-1em}
\end{figure}

\ZTedit{ًWe find a steady increase in publications over time with more than 38 datasets currently available over 35 unique language pairs; however, the majority of languages appear only in a single dataset. 
The research is dominated by three language pairs accounting for $\sim$76\% of all papers: Mandarin-English (zho-eng), Hindi-English (hin-eng), and Arabic-English\footnote{Here, we collate several dialects of Arabic.}  (ara-eng) (see \Cref{fig:lang_dist}).
Mandarin-English is the most studied language pair and is covered by $\sim$55\% of all papers. 
Research on Mandarin-English is partially driven by the availability of large datasets such as SEAME~\cite{lyu2010seame} and the ASRU 2019 challenge~\cite{shi2020asru}. 
The availability of datasets appears to be one of the main drivers of research in this area; 
$\sim$77\%  of papers make use of accessible datasets, while the rest use proprietary or unspecified datasets. For a full listing of languages and datasets, see \Cref{tab:langpair_dataset_breakdown} in the Appendix.
}

\ZTdelete{\Cref{fig:lang_dist} shows the total number of papers per year\ZTdelete{\footnote{While our search query covered papers from 2014 onwards, the first relevant papers related to code-switching in end-to-end ASR appeared in 2018.}}, highlighting the three most commonly found language pairs in the surveyed literature: Mandarin-English (zho-eng), Hindi-English (hin-eng), and Arabic-English (ara-eng). 
Arabic covers several dialectal variants, which we group into the same category.
These three languages account for $\sim$76\% of the papers. 
Mandarin-English, covered in $\sim$55\% of the papers, is the most widely studied pair, driven by the availability of large datasets, such as SEAME and the ASRU 2019 challenge. 
Overall, more languages and datasets are covered over time, with a total of 35 unique pairs covered and 38 unique datasets accessible for research, although most of them appeared only once. 
We only counted ``accessible'' data sets, defined as those with a public link, or it's indicated in papers that they can be shared upon request, or at least used in subsequent literature. 
$\sim$77\% of the papers used such accessible datasets, while the rest used proprietary or unspecified datasets. 
The full listing of languages and datasets is shown in \textcolor{red}{Table x Appendix y}. 
}

Below we briefly summarize frequently used accessible datasets in the literature \ZTedit{grouped by the matrix language. Unless otherwise specified, the embedded language is English.\footnote{We use ISO-3 language codes and full language names interchangeably to represent language pairs. See \Cref{tab:langpair_dataset_breakdown} in the Appendix for a complete mapping.}} See \Cref{tab:dataset_duration_complete} for additional details on all datasets we identify as `accessible'. 
\ZTdelete{For coherence, we a grouping based on the matrix languages, embedded language is assumed to be English unless otherwise specified.}

\subsection{Mandarin}
Datasets for Mandarin code-switched data include: 
TALCS \cite{li22j_interspeech}, ASCEND \cite{lovenia2021ascend}, the ASRU 2019 Code-Switching Challenge dataset \cite{shi2020asru}, and the Southeast Asia Mandarin-English (SEAME) corpus \cite{lyu2010seame}, mainly covering the Mandarin-English pair. 
Among these, SEAME---which contains Mandarin- and Singaporean-accented speech---is the most frequently used ($\sim$24\%)\ZTdelete{, appearing in $\sim$24\% of the papers,} followed by ASRU 2019 ($\sim$11\%). 
Despite being the largest dataset, TALCS is only used in $\sim$4\% of the papers.
\ZTdelete{TALCS, despite being the largest dataset in this category, is used in only about 4\% of the papers. 
Note that the SEAME corpus further divides the development set into two splits: \textit{dev\_man}, which is dominated by Mandarin-accented speech, and \textit{dev\_sge}, which primarily contains Singaporean-accented speech.} 

\subsection{Indic Languages}
Datasets covering Indic languages include MUCS \cite{diwan21_interspeech} and ITTG-HingCos~\cite{ganji2019iitg} which primarily cover Hindi-English (hin-eng) and Bengali-English (ben-eng). 
As the second most studied language group, $\sim$11\% of papers address hin-eng and $\sim$3\% cover ben-eng. 
Of papers studying code-switching with Indic languages, $\sim$38\% rely on the MUCS dataset---which includes monolingual speech in 7 Indic languages as well as CS speech hin-eng and ben-eng language pairs. 
In contrast, ITTG-HingCos focuses solely on the hin-eng pair and captures a range of dialectal variation, as its speakers are sourced from multiple states across India.
\ZTdelete{This category includes datasets such as MUCS \cite{diwan21_interspeech} and ITTG-HingCos \cite{ganji2019iitg}, which primarily cover Hindi-English and Bengali-English language pairs. 
Hindi-English is the second most studied pair in the surveyed literature, appearing in approximately 11\% of the papers, while Bengali-English is covered in only about 3\%. 
Most papers focusing on Hindi or Bengali make use of the MUCS dataset, which is used in around 38\% of those papers, likely due to its large size and its inclusion of both monolingual speech in seven Indic languages as well as code-switched speech. 
In contrast, ITTG-HingCos focuses solely on the Hindi-English pair and captures a range of dialectal variation, as its speakers are sourced from multiple states across India.}

\subsection{Arabic}
Code-switching datasets for Arabic primarily embed English or French, with a few cases of intra-Arabic dialectal CS (i.e., code-switching between Arabic dialects).\footnote{We observe the following Arabic dialects in the literature: Egyptian, Emirati, MSA, Gulf, Jordanian, Mauritanian, Palestinian and Tunisian.}
\ZTedit{We identify 8 datasets for Arabic in the literature---most explicitly designed for CS research---more than any other language.}
\ZTdelete{Arabic is the most resource-rich language in the surveyed work, represented in 8 datasets; most of which are explicitly designed for CS research.} 
Mixat \cite{ali2024mixat} is the largest dataset, covering Emirati Arabic-English, followed by ArzEn \cite{hamed2020arzen} which covers Egyptian Arabic-English code-switching. 

\subsection{African Languages}
The most commonly used dataset for African languages in our survey is the Multilingual Code-switched Soap Opera Speech \cite{van_der_westhuizen_first_2018}, which contains $\sim$3.6 hours of code-switched speech of a total of 14.3 hours. The dataset contains code-switched speech for isiXhosa, isiZulu, Setswana, and Sesotho---four Bantu languages---paired with English, and occasional instances of code-switching between the Bantu languages.\footnote{Bantu is a branch of the Niger-Congo language family spoken in central, eastern, and southern Africa.}

\subsection{Japanese} 
Japanese-English (jpn-eng) is another common language pair in the survey, with about 5\% of papers focusing on it. Interestingly, none of these works use or release a dedicated jpn-eng code-switched dataset. Instead, most rely on synthetic data, typically using the BTEC~\cite{DBLP:journals/ijclclp/TakezawaKMS07} corpus—a parallel Japanese-English dataset for machine translation—to create code-switched segments.
These datasets are proprietary, so no large-scale, publicly available jpn-eng code-switched dataset appears in the surveyed literature.

\section{ASR Modeling Choices}\label{section:asr_modeling}

\ZTedit{We now summarize the findings related to how code-switching has been modeled in end-to-end ASR research.
We present statistics and summarize findings related to: monolingual and multilingual modeling (\Cref{monolingual}), language identification, \Cref{lang_id}), text units (\Cref{input_units}), architectures (Section \ref{sec:architecture}), large pretrained models (Section \ref{sec:pretrained}), decoding schema (\Cref{decoding}), and the use of language models (Section \ref{sec:lm}).}
\ZTdelete{In this section, we summarize the findings related to how code-switching has been modeled in end-to-end ASR research. 
We identified the following design aspects that vary across models: the choice of monolingual or multilingual modeling \ref{monolingual}, the incorporation of language ID \ref{lang_id}, the type of text units \ref{input_units}, the choice of ASR architecture \ref{sec:architecture}, the use of large pretrained models \ref{sec:pretrained}, decoding scheme \ref{decoding}}
\ZTdelete{For each of these, we present statistics from the survey as well as a literature review, summarizing notable research findings that could be insightful for future research.}

\subsection{Monolingual vs. Multilingual Models} \label{monolingual}
\ZTedit{Monolingual modeling has become an important method for optimizing CS performance, even in medium and high-resource settings. 
We find that roughly}
\ZTdelete{Roughly} 60\% of surveyed papers included additional monolingual data for training their models\ZTedit{---half of which include datasets with more than 100 hours of CS speech}. 
\ZTdelete{Roughly half of them include datasets with more than 100 hrs of CS speech.} 
\ZTdelete{This demonstrates that monolingual modeling is an important factor for optimizing CS performance, even in medium- and high-resource settings.} 
Other efforts turned towards multilingual modeling to encode both languages in the latent space. 
More than a third of surveyed paper include multilingual components, where two or more languages are modeled as part of the same ASR architecture. 
These models can be appealing because they capture shared acoustic and lexical patterns across languages\ZTdelete{; moreover, they eliminate}, and they obviate the need for separate language dependent acoustic models and pronunciation dictionaries.

\paragraph{Effect of monolingual training:}
Due to the limited availability of large-scale CS training data, which remains a major challenge in building effective CS ASR systems for most languages, many researchers model each language separately using monolingual data and introduce CS data at a later stage, effectively utilizing transfer learning~\cite{yue_end--end_2019, yang_effective_2024, wang_tri-stage_2024}.
For example, \citet{wang_tri-stage_2024} propose a tri-stage training strategy for a two-pass E2E Mandarin-English ASR system. 
\ZTedit{They use large monolingual Mandarin and English corpora to pretrain two symmetric, language-specific encoders. The pretrained representations are then combined using a feed-forward neural network and the combined system is retained  
using monolingual data, 
followed by fine-tuning on a CS corpus.}
\ZTdelete{First, they pretrain two symmetric, language-specific encoders on large-scale monolingual Mandarin and English corpora. 
Next, a feed-forward neural network is used to combine language-specific representations and transfer the monolingual acoustic attributes to code-switching properties. 
Finally, the proposed framework is fine-tuned on a CS corpus.}

\paragraph{Effect of CS on monolingual performance:} 
\ZTedit{While monolingual modeling can have positive impacts on CS, \citet{Shah2020} argue that the relationship is not necessarily isomorphic. 
They argue that fine-tuning pretrained monolingual models for CS may impair performance on monolingual datasets due to CS data contributing to catastrophic forgetting. 
They propose the Learning Without Forgetting (LWF) mechanism for when the monolingual model is available but its training data is not; and a regularization method when the monolingual training data and model are available. 
LWF applies a knowledge distillation loss to retain the original model's predictions during CS training, which consistently outperforms simple fine-tuning on monolingual and CS test sets; while the regularization method minimizes the KL divergence between the output distribution of the pretrained and fine-tuned models.}
\ZTdelete{One aspect that should be taken into consideration is the potential negative effect on monolingual performance as a result of efforts to optimize models on CS data. 
\citet{Shah2020}  posit that fine-tuning pretrained monolingual models with CS data impairs performance on monolingual datasets due to the ``catastrophic forgetting'' problem, which is exacerbated by increased CS data integration. 
To mitigate this, they propose two strategies. 
1) When the monolingual model is available but the training data is not, they use Learning Without Forgetting (LWF) technique, which applies a knowledge distillation loss to retain the original model’s predictions during CS training. 
They find that LWF consistently outperforms simple fine-tuning on both monolingual and CS test sets. 
2) When both the model and monolingual data are available, they introduce a regularization method that minimizes the KL divergence between the output distributions of the pretrained and fine-tuned models.}

\paragraph{Effect of multilingual training} 
\citet{seki_end--end_2018, seki_end--end_2019} find that multilingual methods may outperform language-dependent models (i.e., models with language-specific ASR modules and a separate LID module) to optimize both CS and monolingual performance. They propose a monolithic multilingual ASR system that jointly performs language identification and transcription, which introduces dynamic language tracking within an utterance to handle intrasentential code-switching. 

\ZTdelete{\citet{seki_end--end_2018} propose a monolithic multilingual ASR system capable of jointly performing language identification and transcription. 
To handle intrasentential code-switching, they introduce dynamic language tracking within an utterance, allowing the model to detect language switches at any point and adapt the output language accordingly during decoding. 
Their analysis shows that this language-independent approach outperforms language-specific models (models containing language dependent ASR modules and a separate language identification module) on evaluation sets containing both intrasentential CS and non-CS utterances.} 

\subsection{Language ID} \label{lang_id}
We find that LID plays a supporting role in many E2E ASR systems for CS in our survey---approximately 33\% of papers incorporate LID in some stage of their training process, either as a pre-processing step to separate the languages~\cite{lu_bi-encoder_2020}, as part of the loss function~\cite{zeng_end--end_2019, shan_end2end, zhao_adapting_2025}, as an auxiliary task~\cite{qiu_towards_2020}, or as part of the predicted output sequence~\cite{seki_end--end_2018, seki_end--end_2019, zhang_rnn-transducer_2020}. 
\ZTdelete{In our survey, 33.6\% of papers incorporated language ID at some stage of their model. 
This can be as a pre-processing step to separate the languages~\cite{lu_bi-encoder_2020}, as part of the loss function~\cite{zeng_end--end_2019, shan_end2end, zhao_adapting_2025}, or as an auxiliary task~\cite{qiu_towards_2020}.  
Other works~\cite[e.g.,][]{seki_end--end_2018, seki_end--end_2019, zhang_rnn-transducer_2020} augment the output vocabulary so the model explicitly predicts the language alongside the transcription, essentially treating LID as an output token.}\ZTedit{
For example, \citet{lu_bi-encoder_2020} propose a bi-encoder transformer network based Mixture-of-Experts (MoE) architecture, which uses an external LID to route language-specific data to the corresponding language-specific encoder to avoid cross-lingual contamination of the encoder modules. 
}
\ZTdelete{\citet{lu_bi-encoder_2020} propose a bi-encoder transformer network based on a Mixture-of-Experts (MoE) architecture. 
Each encoder (expert) is pretrained on monolingual Mandarin and English data, respectively. 
Here, prior LID information is used as a preprocessing step to route language-specific data to the correct encoder, enabling the model to learn language-specific features without cross-language contamination. 
This approach works effectively in this case because the LID information is already known; nonetheless, in scenarios where prior LID information is unavailable, the system’s performance becomes heavily reliant on the accuracy of the LID prediction module. }
\ZTedit{In contrast, }\ZTdelete{\citet{zeng_end--end_2019} add LID loss to a multi-task learning (MTL) setup, combining it with CTC and attention losses in a hybrid CTC/attention framework. 
However, their proposed system does not achieve lower TER than the DNN-HMM baseline.} \citet{shan_end2end} \ZTedit{propose a Multi-Task Learning (MTL) setup where they analyze the impact of LID  loss}\ZTdelete{experiment with applying the LID loss} at different points in the network and find that attention-related components (like the attention output and the attention hidden state) yield the best results compared to adding the loss in decoder hidden state, suggesting that acoustic and alignment information is more useful for LID than the language modeling information in the decoder.

\ZTedit{Other work has experimented with predicting LID at each time step as an auxiliary task}
\ZTdelete{\citet{qiu_towards_2020, shan_end2end} both predict language IDs at each time step as an auxiliary task} alongside character predictions \ZTedit{\cite{qiu_towards_2020,shan_end2end}}. 
\citet{shan_end2end} find that this frame-level LID prediction, implemented via MTL, \ZTdelete{performs better compared to}\ZTedit{outperforms} predicting LID token at the beginning of the sequence. 
\ZTedit{They argue that the latter approach is more appropriate for inter-sentential switching than intra-sentential switching, as it does not capture spontaneous switching within sentences.}
\ZTdelete{They argue that the latter approach, which works well for inter-sentential switching, performs poorly in intra-sentential scenarios, likely because it struggles to model spontaneous, unpredictable switches within a sentence.}

\subsection{Text Units} \label{input_units}
The choice of text units in code-switching ASR, whether shared or distinct at phonetic levels or word/character levels, impacts the model's ability to capture linguistic nuances and transitions between languages. 
A small subset of surveyed papers ($\sim$5\%) used a common label set shared across languages. 
For instance, \citet{seki_end--end_2018, seki_end--end_2019} construct a common character set by taking the union of characters from all involved languages. 
While simple, this approach suffers from \ZTedit{greater confusion between cross-language targets---due to similar acoustic realization of different symbols, and higher computational costs caused by the larger vocabulary}\ZTdelete{two key issues: increased confusability among cross-language targets due to similar acoustic realizations of different symbols, and higher computational cost due to the expanded vocabulary}. 
\ZTedit{Seeking to address this issue, \citet{dhawan_investigating_2020} and \citet{ sreeram_exploration_2020} propose a reduced set of target phones based on acoustic similarity using IITG-HingCos. 
Despite the reduction in the size of the target set, \citet{sreeram_exploration_2020} report an increase in WER which they attribute to ambiguities among homophones.
}
\ZTdelete{\citet{dhawan_investigating_2020, sreeram_exploration_2020} address these issues by proposing a reduced target phone set based on acoustic similarity using the IITG-HingCos corpus. 
While both approaches reduce the target set size, \citet{sreeram_exploration_2020} still report an increase in WER, attributing it to ambiguities among homophones.} 
\ZTedit{\citet{li_bytes_2019} instead propose using Unicode bytes as output units which affords a fixed-size and langauge independent vocabulary, thereby avoiding the need to modify the softmax output layer due to new characters or languages.} 
\ZTdelete{An alternative solution is proposed by \citet{li_bytes_2019}, who move away from language-specific units entirely and propose using Unicode bytes as output units, yielding a fixed-size, language-independent vocabulary which avoids the need to modify the softmax output layer when adding new characters or languages.} 
\ZTedit{Nearly all papers on Mandarin-English CS adopt mixed units, where the former is encoded using characters and the latter using BPE~\cite[e.g.,][]{zhang_rnn-transducer_2020, DBLP:journals/ejasmp/LongWLL21, DBLP:conf/icassp/YanWKJW23}}.
\ZTdelete{Some studies also use subword units, such as BPE \cite{chowdhury2021towards, sailor2021sri, hamed2023benchmarking}; nearly all zho-eng papers adopt mixed units, with Mandarin encoded using characters and English using BPE~\cite{zhang_rnn-transducer_2020, DBLP:journals/ejasmp/LongWLL21, DBLP:conf/icassp/YanWKJW23}. 
Others use simple character sequences.}

\subsection{Architectures}\label{sec:architecture}
\ZTedit{While using CTC is attractive for CS due to the output independence assumption, auto-regressive models with encoder decoder architectures are shown to perform better in \citet{Peng2021}. }
Almost half of the surveyed papers (47\%) employ an encoder-decoder architecture, with RNN/LSTM  \cite{zeng_end--end_2019, 8682824} \ZTedit{being most common before 2020}, and Transformer architectures being more common recently \cite{9054138, 10274649, sailor2021sri, kronis-etal-2024-code}.
Around 17\% of the papers relied on pretrained models, such as Wav2Letter2+ \cite{9413806}, XLS-R \cite{ogunremi2023multilingual, akhi2024transformer}, Wav2Vec2 \cite{10634607}, and most commonly Whisper \cite{ugan-etal-2024-decm, 10627333, alharbi2024leveraging}; these are discussed in more details in the next section.
The remaining papers used CTC-based encoder only models \cite{9688174, tian2022lae}, RNN or Transformer based transducers \cite{9413562, zhang_rnn-transducer_2020}, or Mixture of Experts 
\cite{lu_bi-encoder_2020, 10389662, yang_effective_2024}.  

\subsection{Large Multi-Lingual Pretrained Models}\label{sec:pretrained}

\ZTedit{The use of pretrained language models is another common feature across our sample. 
In particular, $\sim$17\% of papers use Whisper \cite{radford2022robust}, Wav2Vec 2.0 \cite{baevski2020wav2vec}, or XLS-R \cite{conneau2020unsupervised}. 
}
\ZTdelete{In our survey, around 17\% papers reportedly used one of the pretrained multilingual models: Whisper \cite{radford2022robust}, Wav2Vec 2.0 \cite{baevski2020wav2vec}, and XLS-R \cite{conneau2020unsupervised}.} 
\ZTedit{We identify three notable directions in work that use pretrained models: (i) methods that focus on distillation and parameter efficient fine-tuning (PeFT), such as \citet{tseng2024leave} who use a filtering method to distill Whisper Large V2 by 50\% and speed up generation five-fold, while outperforming the teacher model by 30\% in some settings, and \citet{10627333} present Gated Low Rank Adaption, a weight separation-based PeFT method to enable the use of pretrained models on low-spec devices, e.g., mobile phones. 
(ii) Methods that perform multilingual fine-tuning of pretrained models---specifically XLSR---for low-resource languages, e.g., Southern Bantu languages from South Africa \cite{ogunremi2023multilingual}, Kichwa \cite{taguchi-etal-2024-killkan} from the South America and Bangla \cite{akhi2024transformer} from Bangladesh. 
Finally, (iii) simple prompting techniques have been explored. For example, \citet{penagarikano2023semisupervised} show that by concatenating LID tags with the input prompt, Whisper can be used for zero-shot CS detection for Mandarin-English despite not officially supporting code-switching.
}
\ZTdelete{Due to the large size of such pretrained models, some work focuses on distillation and parameter efficient fine-tuning (PeFT) approaches for reducing model size and trainable parameters. 
For instance, \citet{tseng2024leave} propose a data filtering based method to distill Whisper Large V2 to half the size along with a 5X inference speed up while outperforming the teacher model by 30\% in out-of-domain scenario. 
Similarly, \citet{10627333} present a weight separation based PeFt technique called GLoRA (Gated Low Rank Adaptation) enabling pretrained models to be used on low-spec devices like mobile phones. 
\citet{he2023learning} use two language specific adapters for the matrix and embedded language for enhancing performance of Wav2Vec 2.0. 
Multilingual pretraining of such models especially for XLSR enables their adoption towards low-resource languages like Southern Bantu languages from South Africa \cite{ogunremi2023multilingual}, Kichwa \cite{taguchi-etal-2024-killkan} from the South America and Bangla \cite{akhi2024transformer} from Bangladesh. 
All of these papers are able to achieve good results by a simple fine-tuning of the XLSR model. 
Additionally, simple prompting technique like simple concatenation of the language ID tokens in the input prompt in the case of Whisper leads to the emergence of zero-shot CS ability for Mandarin-English despite the fact that Whisper does not support code-switching out of the box \cite{penagarikano2023semisupervised}.} 

\subsection{Decoding Strategies} \label{decoding}
Most papers do not explicitly mention the decoding strategy. 
As greedy decoding is the simplest strategy and is commonly used in end-to-end models, we assume that most papers follow this strategy (only seven papers explicitly mention greedy decoding). 
Around 30\% of papers explicitly mention a strategy other than greedy decoding\ZTedit{, approximately 70\% of which used some form of beam search.} 
\ZTdelete{Among these, almost 70\% used some form of beam search.} 
Other papers 
have used multi-graph decoding~\cite{yue_end--end_2019, Ali2021}, or other techniques. 
For instance, \citet{yue_end--end_2019} introduce a multi-graph decoding strategy to address data imbalance between high-resourced and low-resourced languages. 
They 
construct parallel Weighted Finite-State Transducer (WFST) search graphs, one for each monolingual language and one for the bilingual setup (Frisian-Dutch), while sharing the acoustic and lexicon models, which allows the decoder to dynamically select the most appropriate path during inference. 
However, this setup does not support intra-sentence switching, as decoding is limited to a fixed monolingual or bilingual path. 
In contrast, \citet{Ali2021} propose a similar multi-graph decoding approach but incorporate a Kleene-closure, enabling transitions between language graphs and allowing intra-sentence switching during decoding.
\subsection{Language Model}\label{sec:lm} 
Around 45\% papers \ZTdelete{indicate that they use a language model (LM)}\ZTedit{report using a language model (LM)} in their ASR pipeline to improve decoding accuracy\ZTdelete{, either} through shallow fusion or rescoring. 
\ZTedit{While 20\%}\ZTdelete{A fifth of these papers} use some form of n-gram LMs \cite{tian2022lae, 10389662, srivastava2018homophone}, the \ZTdelete{rest}\ZTedit{remainder} use RNN-based LMs \cite{yue_end--end_2019, Sharma2020, zhou_multi-encoder-decoder_2020} or transformer-based LMs \cite{liu_reducing_2023, 10448335, 10389798}. \citet{li_improving_2020} experiment with \ZTedit{a word-based LM (one-layer LSTM) and a subword-based LM (two-layer RNN) for Mandarin-English code-switched ASR. 
The models are trained on natural and synthetic ASR and integrated into the ASR using shallow fusion.}
\ZTdelete{two language model architectures \ZTdelete{to improve ASR performance on a}\ZTedit{for} Mandarin-English code-switching\ZTdelete{ task}: a word-based LM (one-layer LSTM with 650 units) and a subword-based LM (two-layer RNN with 650 units). 
Both models are trained on a mix of real and artificially generated CS text and integrated into the ASR system via shallow fusion.} 
\ZTedit{They find that the subword-based LM trained on SEAME and augmented with CycleGAN-generated \cite{Zhu_2017_ICCV} text achieves the highest performance.}
\ZTdelete{The subword-based LM trained on SEAME data augmented with CycleGAN-generated~\cite{Zhu_2017_ICCV} CS sentences achieves the best performance, outperforming the word-based LM.} 
\citet{chen-rescoring} investigate \ZTdelete{LLM's like}\ZTedit{the capability of } 
T5~\cite{t5}, MT5~\cite{mt5} and PaLM\ZTdelete{'s}~\cite{palm} \ZTdelete{capability} to rescore hypotheses generated by a \ZTdelete{first-pass} ASR model for long-form speech recognition in US English and Indian English. 
The LLMs compute log-likelihoods of ASR outputs and improve performance through rescoring with optional segment-level context. 
The paper finds that MT5 rescoring---when fine-tuned on CS data---improves WER over neural and maximum entropy based baselines. 

\section{Training \& Evaluation} \label{section:training}

\subsection{Data Augmentation} \label{data_aug} 
\ZTedit{Code-switching is a low-resource setting due to the small size of datasets for code-switching and the limited coverage of language pairs.}
\ZTdelete{Code-switching is considered a low-resource setting since most models are trained with monolingual data, with a few small CS corpora existing for a limited subset of languages. 
Therefore, data augmentation might be necessary to improve performance.} 
\ZTedit{Unsurprisingly, }
\ZTdelete{In our survey, around }\ZTedit{$\sim$}30\% of \ZTdelete{the} papers explicitly mention data augmentation in their methodology. 
SpecAugment \cite{specaugment}, speed perturbation, and \ZTdelete{text-to-speech}\ZTedit{TTS} synthesis \cite{Sharma2020} \ZTdelete{are}\ZTedit{constitute} the most commonly used data augmentation techniques. 

While SpecAugment and speed perturbation remain the main techniques used, some methods have been developed specifically for CS speech. 
\citet{Sharma2020} synthesize Hindi-English CS speech and apply Mixup regularization \cite{mixup} to bridge the distribution gap between synthetic and real data. Results on proprietary data show reductions up to 5\% absolute WER.
\ZTedit{\citet{Du2020} find mixed results in an exploration of TTS-based augmentation. }
\ZTdelete{TTS-based augmentation is also explored by \citet{Du2020} with mixed results.} 
They experiment with audio splicing, random noun/verb translation, and random English word insertion into Mandarin sentences followed by TTS. 
While the combination of all three methods along with SpecAugment yielded the best overall performance, SpecAugment still outperformed them individually. 
\ZTdelete{\citet{DBLP:journals/ijclclp/TakezawaKMS07} similarly rely on English word insertion into Japanese.\ZC{Removing this, I can't find anything about word insertion in the paper.}}
\citet{Liang2022} compare numerous data augmentation techniques, namely pitch shifting, speed perturbation, audio codec augmentation, SpecAugment, and a TTS approach combining transcripts from one dataset and style IDs from another. 
Experiments on a Conformer favor TTS and speed perturbation, while audio codec and pitch shifting can have an adverse effect on performance. 
Other methods \cite{chi23_interspeech, speech-collage} explore CS speech generation from monolingual data, using grid beam search \cite{grid-beam} and audio splicing with energy normalization.

\subsection{Translation} \label{translation}

\ZTedit{17 papers \ZTdelete{in our survey} use translation in their methodology to augment their language model data~\cite{penagarikano2023semisupervised}, to improve model evaluation \cite{taguchi-etal-2024-killkan, polywer}, or to used TTS synthesis to augment their speech data \cite{10096317, tazakka2024indonesian}.
}
\ZTdelete{In our survey, 17 papers explicitly mentioned translation in their methodology. Some used translation to augment their language model data \cite{penagarikano2023semisupervised}, and others to improve model evaluation \cite{taguchi-etal-2024-killkan, polywer}. Some used translation alongside text-to-speech synthesis to augment their speech data \cite{10096317, tazakka2024indonesian}.} 

\ZTedit{For example, \citet{10096317} use machine translation to generate a parallel dataset of sentences, then perform word alignment and substitution to ensure the synthetic data mirrors statistics of CS in the natural data. 
They convert their synthesized text into speech using a multilingual TTS system to augment their training data. 
Their method obtains a 16\% relative reduction in error rate for Mandarin-English. 
\citet{tazakka2024indonesian} take a similar approach to perform semi-supervised training for Indonesian-English.
}
\ZTdelete{
\citet{10096317}, for instance, generate synthetic CS sentences using machine translation to generate parallel sentences, followed by word alignment and substitution, ensuring that CS statistics are consistent with training data. 
A multilingual TTS system is used to convert these sentences into speech, which is used to augment training data, showing 16\% relative reductions in error rates for Mandarin-English. 
A similar approach was followed for Indonisian-English ASR in \citet{tazakka2024indonesian}, using machine speech chain \cite{9020132} for semi-supervised training.
} 

\subsection{Zero-Shot Evaluation} 
\label{zero_shot_eval}
Given the scarcity of CS speech data and the prevalence of multilingual ASR systems, zero-shot evaluation appears to be an attractive choice for CS ASR. 
\ZTedit{While using large multilingual pretrained models, such as Whisper presents as an obvious choice, Whisper is also only trained with monolingual data. 
However, \citet{peng2023prompting} find that concatenating the language tokens for the languages in a CS scenario improves Whisper performance, despite it not being explicitly trained to receive such combined tokens.}
\ZTdelete{Using large multilingual pretrained models like Whisper is one obvious choice. 
However, Whisper is also trained with monolingual language IDs. 
\citet{peng2023prompting} find that concatenating the two language tokens improves Whisper performance, despite it not being explicitly trained to receive such combined tokens.}
\citet{ugan-etal-2024-decm} conduct a zero-shot evaluation of different ASR models on their German-English DECM dataset. 
They use the massively multilingual speech dataset \cite[MMS,][]{mms}
\ZTdelete{MMS} and two large Whisper models, one of which is used with a German decoding prefix (WhisperDe). 
Whisper performs best overall with WhisperDe close behind, while MMS has almost a 12\% higher WER. 
Further analysis suggests that WhisperDe’s lower overall performance stems from its superior German monolingual performance, rather than CS in particular. 
\citet{zhou-knnctc-dual} adapt kNN-CTC \cite{zhou-knnctc} to enable zero-shot Chinese-English ASR through the use of dual monolingual datastores of labeled examples. 
This setup ensures that retrieved examples with similar audio frames come from the appropriate language datastore during decoding, with the help of a gating mechanism. 
This approach outperforms CTC fine-tuning and even small Whisper variants \cite{radford2022robust, peng2023prompting} \ZTedit{that are} almost double the parameter size. 
In \citet{DBLP:conf/icassp/YanWKJW23}, two monolingual modules transcribe all segments using their respective vocabularies, resulting in transliterations for foreign segments. 
This delays the CS boundary decision to a textual bilingual module that generates a bilingual sequence from the previous outputs, thereby mitigating error-propagation compared to earlier methods \cite{tian2022lae, song22e_interspeech}.

\begin{table*}[!ht]
\begin{center}
\scriptsize
\begin{tabular}{*{14}{c}}
\toprule
\rotatebox{90}{\scriptsize Dataset} & \rotatebox{90}{\scriptsize Best Model} & \rotatebox{90}{Year} & \rotatebox{90}{\scriptsize Monolingual} & \rotatebox{90}{\scriptsize Multilingual} & \rotatebox{90}{LM} & \rotatebox{90}{LID} & \rotatebox{90}{\scriptsize Translation} & \rotatebox{90}{\scriptsize Augmentation} & \rotatebox{90}{\scriptsize Zero-shot} & \rotatebox{90}{Architecture} & \rotatebox{90}{Languages} & \rotatebox{90}{Metric} & \rotatebox{90}{Best Result} \\
\midrule
SEAME & \cite{aditya2024attention}& 2024 & \checkmark &  &  &  &  &  &  & Whisper & zho-eng & MER & 14.2 \\
 &  &    &  &  &  &  &  &  &  &  & sgp-eng &  & 20.8 \\
ASRU & \cite{wang2023language}& 2023 & \checkmark &  &  & \checkmark &  & \checkmark &  & MoE & zho-eng & MER & 8.2 \\
TALCS & \cite{wang_tri-stage_2024} &2024 & \checkmark & \checkmark &  &  &  &  &  & Enc & zho-eng & MER & 6.17 \\
ASCEND & \cite{tseng2024leave} &2024 & \checkmark &  &  &  &  & \checkmark &  & Whisper & zho-eng & MER & 17.86 \\
MUCS & \cite{kumar2021dual}& 2021 &  & \checkmark & \checkmark & \checkmark & \checkmark &  &  & Enc-Dec & hin-eng  & WER & 22.0 \\
      & &     &  &  &  &  &  &  &  &  & ben-eng & WER  & 27.8 \\
ArzEn &\cite{hamed2022investigations} & 2022 & \checkmark & \checkmark & \checkmark &  &  & \checkmark &  & Non E2E & ara-eng  & WER & 32.1 \\

Mixat &\cite{polywer} &2024 &  &  &  &  &  \checkmark &  &  & Whisper & ara-eng & WER & 24.8 \\

Miami Bangor& \cite{hillah2024te}& 2024 &  & \checkmark &  & \checkmark &  &  &  & Whisper & spa-eng & WER & 48.38 \\

ESCWA& \cite{chowdhury2021towards}& 2021 &  & \checkmark &  &  &  &  &  & Enc-Dec & ara-eng & WER & 37.7 \\

\bottomrule
\end{tabular}
\end{center}
\caption{Best performing models on popular datasets.} \label{tab:sota}
\end{table*}

\newpage
\subsection{Evaluation Metrics}
\ZTedit{We now turn to discussing notable metrics used in the surveyed work (see \Cref{tab:metrics} in the Appendix for a full listing of metrics).}
\ZTdelete{Table \ref{tab:metrics} lists all the metrics found in our survey, with a brief description, and a list of languages for which the metric was proposed/used. Below, we categorize and summarize notable examples.} 

\paragraph{Standard Metrics:} The standard metric for evaluating ASR accuracy is the Word Error Rate (WER). 
WER computes the edit distance between a reference and prediction 
and divides the total number of substitutions $(S)$, insertions $(I)$, and deletions $(D)$ by the total number of words in the reference $(N)$: \( \mathrm{WER} = (S + I + D)/N \).

\noindent Character Error Rate (CER) offers a higher level of granularity by employing the same principle of WER but on a character-level.
It is particularly useful for languages that lack word boundaries such as Chinese, Japanese and Thai.

\paragraph{Mixed Metrics:} 
For code-switching, \ZTdelete{several} other metrics \ZTedit{such as  Mixed Error Rate (MER) and Token Error Rate (TER) \cite{zhou-knnctc-dual,Shen2022}} have been proposed due to the nature of code-switching that often involves different writing systems, e.g., where one language employs word boundaries but another does not (such as Japanese-English). 
\ZTedit{ 
MER combines WER for word-based language segments and CER for character-based segments within the same transcription, }
\ZTdelete{In cases where both types of languages are used (e.g., Japanese-English), Mixed Error Rate (MER) and Token Error Rate (TER) are popular choices \cite{zhou-knnctc-dual, Shen2022}. 
MER combines WER for word-based language segments and CER for character-based ones within the same transcription,} while TER evaluates prediction at the token-level, be it a character for a language like Chinese or a sub-word unit for languages like English\ZTdelete{ \cite{zhou_multi-encoder-decoder_2020}}.

\paragraph{Transliteration-Based Metrics:} 
\ZTedit{Other metrics first transliterate to a single writing system, then compute performance to address the ambiguous distinction between code-switching and loan words in cases where the matrix and embedded languages use different scripts and predictions may be unfairly penalized for being written in the wrong script.
For example, Transliteration-Optimized WER \cite[toWER,][]{emond2018transliteration} first transliterates using a weighted finite state transducer then computes WER on the ``transliterated space'', and the Pronunciation-Optimized Word Error Rate \cite[poWER,][]{srivastava2018homophone} process the reference and prediction with a grapheme-to-phoneme conversion for each word to ensure that predicted words that are faithful to the audio are not penalized. \citet[polyWER,][]{polywer} instead propose a modification to the edit distance algorithm to enable multi-reference evaluation. 
They apply PolyWER across two dimensions: transliteration with a CER threshold for allowed variations in orthography, or 
translation with a cosine similarity threshold to allow synonyms. 
}
\ZTdelete{
\citet{emond2018transliteration} introduced Transliteration-Optimized WER (toWER) to address the ambiguous distinction between code-switching and loan words in cases where the matrix and embedded languages use different scripts. This ambiguity often causes predictions to be unfairly penalized for being written in a script other than the one used in the reference. They achieve that by mapping all the transcriptions to a single writing system using a weighted finite state transducer, before performing WER on this shared ``transliterated space''. 
Transliterated WER (TW) \cite{chowdhury2021towards} follows the same principle using the transliteration system proposed by \citet{dalvi-qcri}
\citet{srivastava2018homophone} approach the same problem with Pronunciation-Optimized Word Error Rate (poWER), where both prediction and reference sentences go through a grapheme-to-phoneme (g2p) conversion
for each word. This ensures that predicted words that are faithful to the audio are not penalyzed regardless of script. 
PolyWER \cite{polywer} modifies the edit distance algorithm to enable multi-reference evaluation. They do that across two possible dimensions: transliteration with a CER threshold for allowed variations in orthography, and optionally translation with a cosine similarity threshold to allow synonyms. 
}

\section{Best Performing Models}\label{sec:sota}

We summarize the modeling choices corresponding to the state-of-the-art approaches on popular datasets in Table \ref{tab:sota}. Note that there is no single methodology shared across all datasets. Roughly half use monolingual modeling, and half use multi-lingual modeling, with only two SOTA systems utilizing both approaches. The use of LM, LID, and data augmentation is relatively small, in spite of efforts showing the potential of these approaches. None of the SOTA models rely on zero-shot evaluation, which underscores the importance of dedicated CS training.  The choice of architecture also varies, but Encoder-Decoder architectures (including Whisper) is most common. In one of the datasets (ArzEn), the explored E2E models do not even outperform the sequential baseline, which is a hybrid CNN-TDNN model. 

\section{Discussion: Challenges \& Opportunities}\label{section:challenges}

Research in code-switching is undergoing rapid methodological innovation and resource development. 
Here, we present a discussion of challenges and opportunities as drawn from our analysis of the surveyed literature. 

\paragraph{Data Scarcity:}
Research in this area is largely driven by data availability, which has led to a stronger focus on higher-resourced languages. For instance, Mandarin-English code-switching has attracted the most attention in end-to-end ASR due to the pre-existing datasets and well-defined benchmarks.
Moreover, while there is an upward trend in the publication of new resources, they remain concentrated around a small subset of languages, thereby privileging some languages while leaving many others under-resourced and under-studied. 
Adding to the challenge of resource availability is that a large proportion of datasets used in the surveyed literature are proprietary, and therefore are not available to the wider research community, e.g., for further studies or replicating results.
The development of publicly available training and evaluation data thus presents an opportunity for researchers to help steer the future of research towards under-represented languages.

\paragraph{Disparities in coverage: }
While ASR research is moving towards inclusive and fair representation with coverage of hundreds of languages, research in code-switching in ASR remains confined to a small subset of languages. Moreover, dialectal variations are only partially represented for the Arabic and Indic language families. Considering the colloquial nature of code switching, dialectal variations also need to be taken into account, with a focus on marginalized communities who often need to code-switch in order to utilize such systems. As ASR technology continues to be deployed in critical domains such as accessibility, healthcare, and education, this disparity in coverage risks exacerbating existing inequalities. 

\paragraph{Evaluation:}

The recent introduction of new evaluation metrics provides further avenues for precise evaluation of end-to-end ASR systems for code-switching, but they also underscore the existing conceptual challenges in ASR evaluation and the shortcomings of existing metrics. 
\ZTedit{These challenges arise due to differences between scripts (e.g., different levels of tokenization) and norms (e.g., standardization of spelling and transliteration).}
This calls for additional efforts to examine current evaluation practices and propose metrics that are \ZTedit{valid (i.e., consistent, interpretable, and meaningful) \cite{jacobs2021measurement,delobelle_metrics_2024}} for a \ZTedit{grounded} assessment of progress.

\paragraph{Systematic studies \& Reproducibility:}

The rapid methodological innovation in the field has led to a lack of insight into (i) the robustness of proposed methods, (ii) their applicability across language pairs, and (iii) the interaction of different methods, e.g., the impact \ZTedit{of} data augmentation methods on different architectures or language pairs. 
Furthermore, with the exception of \ZTdelete{few}\ZTedit{of a small set of} languages, the efforts appear sporadic and lack a consistent evaluation framework, benchmarking, and continuity to compare proposed methods with past efforts.  
As a result, positive findings from earlier studies have often not been replicated on newer datasets, leading to performance gaps \ZTedit{and uncertainty about the robustness and validity of prior work}.
These gaps present opportunities for future research, such as the development of benchmarks to standardize experimental methodology and studies focused on replication and comparison.

\paragraph{Data Augmentation:}

Data augmentation lies at the intersection of resource and methodological challenges. 
In the surveyed literature, data augmentation, e.g., through code-switched speech synthesis and audio perturbation, has been frequently employed as a redress to concerns of data sparsity. 
While increasing data resources for a particular language may mitigate the data sparsity issue, data augmentation can still prove useful for other purposes, e.g., to ensure model robustness and generalization. Furthermore, the availability of monolingual data will likely continue to outpace that of code-switching resources, which remain limited in part due to the challenges of annotation. 
The field could benefit from further innovation in data augmentation techniques and comprehensive studies in the utility of data augmentation across different language pairs and settings.

\section{Conclusion}

Research on code-switching in end-to-end ASR is experiencing a boom with an increasing number of papers published in recent years. 
In this paper, we conducted an extensive literature review of 127 papers published on the topic of E2E ASR for code-switched speech. 
By annotating and analyzing these papers, we sought to identify and explicate developments and trends in this body of research across languages and datasets, modeling choices, and training and evaluation methods, and presented a discussion of existing and potential challenges identified through our literature review. We note a significant bias towards language pairs with established datasets and benchmarks, underscoring the importance of resources and standardization in driving research. Through this study, we hope to encourage future research towards more inclusive and replicable research.

\bibliography{refs}
\bibliographystyle{acl}

\appendix
\clearpage
\onecolumn
\section{Lists of Datasets, Languages, Metrics and Papers}

\begin{table*}[!ht]
\centering
\small
\renewcommand{\arraystretch}{1.1}
\resizebox{0.98\textwidth}{!}{%
\begin{tabular}{llr|p{0.65\textwidth}}
\toprule
\textbf{Language Pair} & \textbf{ISO-3 Code} & \textbf{\# Papers} & \textbf{Dataset Breakdown} \\
\midrule
Mandarin–English & zho-eng & 70 & \dsentry{seamecolor}{32} \dsentry{asru2019color}{15} \dsentry{talcscorpuscolor}{6} \dsentry{ascendcolor}{5} \dsentry{magicdataramccolor}{2}  \dsentry{asru2021color}{1} \dsentry{aicubescompetitiondatasetcolor}{1} \dsentry{iscslp2022magichubcodeswitchingasrchallengedatasetcolor}{1} \dsentry{ntutab01color}{1} \dsentry{proprietarydatacolor}{11} \\

Hindi–English & hin-eng & 15 & \dsentry{mucscolor}{5} \dsentry{iitghingcoscolor}{3} \dsentry{iiithhecmcolor}{1} \dsentry{vitbhebiccolor}{1}
\dsentry{proprietarydatacolor}{5} \\

Arabic–English & ara-eng & 12 & \dsentry{arzncolor}{3} \dsentry{escwacolor}{2} \dsentry{mixatcolor}{2} \dsentry{qasr.cscolor}{1} \dsentry{haccolor}{1} \dsentry{tunswitchcolor}{1} \dsentry{zaebucspokencolor}{1} \dsentry{casablancacolor}{1} \dsentry{saudilangcorpus(scc)(evalonly)color}{1} \\
Japanese–English & jpn-eng & 7 & \dsentry{proprietarydatacolor}{7} \\
Bengali–English & ben-eng & 5 & \dsentry{mucscolor}{3} \dsentry{proprietarydatacolor}{2} \\
German–English & deu-eng & 4 & \dsentry{germanspokenwikipediacorpuscolor}{1} \dsentry{decmcolor}{1} \dsentry{proprietarydatacolor}{2} \\
Korean–English & kor-eng & 3 & \dsentry{medreccolor}{1}  \dsentry{kecscolor}{1}  \dsentry{proprietarydatacolor}{1} \\
Spanish–English & spa-eng & 3 & \dsentry{bangormiamicolor}{2} 
\dsentry{proprietarydatacolor}{1} \\
French–English & fra-eng & 3 & \dsentry{proprietarydatacolor}{3} \\
Italian–English & ita-eng & 2 & \dsentry{proprietarydatacolor}{2} \\
Dutch–English & nld-eng & 2 & \dsentry{proprietarydatacolor}{2} \\
Portuguese–English & por-eng & 2 & \dsentry{proprietarydatacolor}{2} \\
Russian–English & rus-eng & 2 & \dsentry{proprietarydatacolor}{2} \\
Thai–English & tha-eng & 2 & \dsentry{lotusbicolor}{1} \dsentry{proprietarydatacolor}{1} \\
Arabic–French & ara-fra & 2 & \dsentry{escwacolor}{1} \dsentry{qasr.cscolor}{1} \dsentry{tunswitchcolor}{1} \\
Indonesian–English & ind-eng & 2 & \dsentry{hartantocscolor}{2} \\
isiZulu–English & zul-eng & 2 & \dsentry{southafricansoapoperascolor}{2} \\
isiXhosa–English & xho-eng & 2 & \dsentry{southafricansoapoperascolor}{2} \\
Sesotho–English & sot-eng & 2 & \dsentry{southafricansoapoperascolor}{2} \\
Setswana–English & tsn-eng & 2 & \dsentry{southafricansoapoperascolor}{2} \\
Sepedi–English & nso-eng & 2 & \dsentry{southafricansoapoperascolor}{2} \dsentry{twosepedienglishcsspeechcorpus}{1} \\
Frisian–Dutch & fry-nld & 1 & FAME! \\
French–Mandarin & fra-zho & 1 & \dsentry{proprietarydatacolor}{1} \\
Japanese–Mandarin & jpn-zho & 1 & \dsentry{proprietarydatacolor}{1} \\
Tamil–English & tam-eng & 1 & \dsentry{proprietarydatacolor}{1} \\
Cantonese–English & yue-eng & 1 & \dsentry{proprietarydatacolor}{1} \\
Marathi–English & mar-eng & 1 & \dsentry{proprietarydatacolor}{1} \\
Kazakh–English & kaz-eng & 1 & KSC2 \\
Basque–Spanish & eus-spa & 1 & Bilignual Basque-Spanish Dataset \\
Manipuri–English & mni-eng & 1 & MECOS \\
Quechua–Spanish & que-spa & 1 & Killkan \\
Urdu–English & urd-eng & 1 & Roman Urdu Code-Mixed Dataset \\
Latvian–English & lav-eng & 1 & \dsentry{proprietarydatacolor}{1} \\

\bottomrule
\end{tabular}
}
\caption{Number of papers per language pair and dataset usage breakdown.}
\label{tab:langpair_dataset_breakdown}
\vspace{0.5em}
\textbf{Dataset Legend:} \\

\noindent
\dsbox{seamecolor} SEAME \quad
\dsbox{asru2019color} ASRU 2019 \quad
\dsbox{talcscorpuscolor} TALCS \quad
\dsbox{ascendcolor} ASCEND \quad
\dsbox{magicdataramccolor} MagicData-RAMC \quad
\dsbox{asru2021color} ASRU 2021 \quad

\dsbox{aicubescompetitiondatasetcolor} AICUBES Competition Dataset \quad
\dsbox{iscslp2022magichubcodeswitchingasrchallengedatasetcolor} ISCSLP 2022 CSASR challenge \quad
\dsbox{ntutab01color} NTUT-AB01 \quad

\dsbox{mucscolor} MUCS \quad
\dsbox{iitghingcoscolor} IITG-HingCoS \quad
\dsbox{iiithhecmcolor} IIITH-HE-CM \quad
\dsbox{vitbhebiccolor} VITB-HEBiC \quad

\dsbox{arzncolor} ArzEn \quad
\dsbox{escwacolor} ESCWA \quad
\dsbox{mixatcolor} Mixat \quad
\dsbox{qasr.cscolor} QASR.CS \quad
\dsbox{haccolor} HAC \quad
\dsbox{tunswitchcolor} TunSwitch \quad
\dsbox{zaebucspokencolor} ZAEBUC-Spoken \quad
\dsbox{casablancacolor} Casablanca \quad

\dsbox{saudilangcorpus(scc)(evalonly)color} Saudilang Code-switch Corpus (SCC) \quad

\dsbox{germanspokenwikipediacorpuscolor} German Spoken Wikipedia Corpus \quad
\dsbox{decmcolor} DECM \quad
\dsbox{medreccolor} MEDREC \quad
\dsbox{kecscolor} KECS \quad
\dsbox{bangormiamicolor} Bangor Miami \quad
\dsbox{lotusbicolor} LOTUS-BI \quad

\dsbox{hartantocscolor} Hartanto CS \quad
\dsbox{southafricansoapoperascolor} South African Soap Operas \quad
\dsbox{twosepedienglishcsspeechcorpus} Two Sepedi-English Code-Switched Speech Corpora \quad
\dsbox{proprietarydatacolor} Proprietary Data

\end{table*}

\begin{table}[http!]
  \centering
  \scalebox{0.6}{
  \begin{tabular}{lll}
    \toprule
    \textbf{Metric} 
      & \textbf{Languages} 
      & \textbf{Description} \\
    \midrule
    WER
      & \texttt{spa}, \texttt{kor}, \texttt{hin}, \texttt{fry-nld}, \texttt{nld}, \texttt{cmn}, \texttt{tha}
      & Word‐level ED \\
    \addlinespace
    CER
      & \texttt{jpn}, \texttt{cmn}, \texttt{deu}, \texttt{spa}, \texttt{fra}, \texttt{ita}, \texttt{nld}, \texttt{rus}, \texttt{por}
      & Character‐level ED \\
    \addlinespace
    TER
      & \texttt{jpn}, \texttt{cmn}
      & Token‐level ED \\
    \addlinespace
    poWER
      & \texttt{hin}
      & WER + g2p conversion \\
    \addlinespace
    toWER
      & \texttt{hin}, \texttt{ben}
      & WER + WFST transliteration \\
    \addlinespace
    MER
      & \texttt{cmn}
      & WER (segmental) + CER (syllabic) \\
    \addlinespace
    PPL
      & \texttt{cmn}
      & Token‐level perplexity \\
    \addlinespace
    WpER
      & \texttt{jpn}
      & Wordpiece ED \\
    \addlinespace
    UER
      & \texttt{kor}
      & Unit‐level ED (e.g.\ jamo, syllable) \\
    \addlinespace
    SER
      & \texttt{kor}
      & Sentence‐level ED \\
    \addlinespace
    LER
      & \texttt{cmn}, \texttt{eng}, \texttt{jpn}, \texttt{deu}, \texttt{spa}, \texttt{fra}, \texttt{ita}, \texttt{nld}, \texttt{rus}, \texttt{por}
      & LID error rate \\
    \bottomrule
  \end{tabular}
  }
  \caption{%
    Evaluation metrics and the languages they were employed for in our survey. 
    Embedded language is English unless specified otherwise. 
    ED: Edit distance.
  }
  \label{tab:metrics}
\end{table}

\clearpage
\scriptsize
\begin{longtable}{llllll}
\toprule
\textbf{DOI} & \textbf{Year} & \textbf{Type} & \textbf{Dataset} & \textbf{Languages} \\
\midrule
\href{https://www.isca-archive.org/interspeech_2019/zeng19_interspeech.html}{10.21437/interspeech.2019-1429} & 2018 & Modeling & \dsbox{seamecolor} SEAME & zho-eng \\
\href{https://ieeexplore.ieee.org/document/8682674/}{10.1109/ICASSP.2019.8682674} & 2018 & Modeling & \dsbox{proprietarydatacolor} Proprietary Data & jpn-eng \\
\href{https://www.isca-archive.org/interspeech_2018/srivastava18_interspeech.html}{10.21437/Interspeech.2018-1171} & 2018 & Modeling & \dsbox{proprietarydatacolor} Proprietary Data & hin-eng \\
\href{https://ieeexplore.ieee.org/document/8693044/}{10.1109/ICSDA.2018.8693044} & 2018 & Dataset & \dsbox{proprietarydatacolor} Proprietary Data & jpn-eng \\
\href{https://ieeexplore.ieee.org/document/8639699/}{10.1109/SLT.2018.8639699} & 2018 & Evaluation Metric & \dsbox{proprietarydatacolor} Proprietary Data & hin-eng \\
\href{https://ieeexplore.ieee.org/document/8462180/}{10.1109/ICASSP.2018.8462180} & 2018 & Modeling & \dsbox{proprietarydatacolor} Proprietary Data & zho-eng, jpn-eng, deu-eng, \\ 
& & & & spa-eng, fra-eng, ita-eng, \\
& & & & nld-eng, rus-eng, por-eng \\
\href{https://www.isca-archive.org/sltu_2018/rambabu18_sltu.html}{10.21437/SLTU.2018-23} & 2018 & Dataset & \dsbox{iiithhecmcolor} IIITH-HE-CM & hin-eng \\
\href{https://ieeexplore.ieee.org/document/9004035/}{10.1109/ASRU46091.2019.9004035} & 2019 & Modeling & FAME & fry-nld \\
\href{https://ieeexplore.ieee.org/document/8682824/}{10.1109/ICASSP.2019.8682824} & 2019 & Other & \dsbox{proprietarydatacolor} Proprietary Data & ben-eng \\
\href{https://www.isca-archive.org/interspeech_2019/khassanov19_interspeech.html}{10.21437/interspeech.2019-1867} & 2019 & Modeling & \dsbox{seamecolor}  SEAME & zho-eng \\
\href{https://ieeexplore.ieee.org/document/9003926/}{10.1109/ASRU46091.2019.9003926} & 2019 & Modeling & \dsbox{proprietarydatacolor} Proprietary Data & jpn-eng, jpn-zho, zho-eng, \\ 
& & & & fra-eng, fra-zho \\
\href{https://ieeexplore.ieee.org/document/9037688/}{10.1109/IALP48816.2019.9037688} & 2019 & Modeling & \dsbox{seamecolor}  SEAME & zho-eng \\
\href{https://www.aclweb.org/anthology/K19-1026}{10.18653/v1/K19-1026} & 2019 & Data Augmentation & \dsbox{seamecolor}  SEAME & zho-eng \\
\href{https://ieeexplore.ieee.org/document/9060847/}{10.1109/O-COCOSDA46868.2019.9060847} & 2019 & Modeling & \dsbox{proprietarydatacolor} Proprietary Data & jpn-eng \\
\href{https://ieeexplore.ieee.org/document/8683223/}{10.1109/ICASSP.2019.8683223} & 2019 & Modeling & \dsbox{proprietarydatacolor} Proprietary Data & zho-eng \\
\href{https://ieeexplore.ieee.org/document/9056083/}{10.1109/NCC48643.2020.9056083} & 2019 & Modeling & \dsbox{iitghingcoscolor} IITG-HingCoS & hin-eng \\
\href{https://ieeexplore.ieee.org/document/9041195/}{10.1109/O-COCOSDA46868.2019.9041195} & 2019 & Dataset & \dsbox{lotusbicolor} LOTUS-BI & tha-eng \\
\href{https://linkinghub.elsevier.com/retrieve/pii/S0167639318304217}{10.1016/J.SPECOM.2019.04.007} & 2019 & Dataset & \dsbox{iitghingcoscolor} IITG-HingCoS & hin-eng \\
\href{https://www.isca-archive.org/interspeech_2020/wang20k_interspeech.html}{10.21437/interspeech.2020-2440} & 2019 & Modeling & \dsbox{medreccolor} MEDREC & kor-eng \\
\href{https://ieeexplore.ieee.org/document/8682850/}{10.1109/ICASSP.2019.8682850} & 2019 & Modeling & \dsbox{proprietarydatacolor} Proprietary Data & zho-eng \\
\href{https://www.isca-archive.org/interspeech_2019/seki19_interspeech.html}{10.21437/INTERSPEECH.2019-3038} & 2019 & Modeling & \dsbox{proprietarydatacolor} Proprietary Data & jpn-eng, zho-eng, deu-eng, \\ 
& & & & fra-eng, ita-eng, nld-eng, \\
& & & & por-eng, rus-eng \\
\href{https://www.aclweb.org/anthology/2020.acl-main.348}{10.18653/v1/2020.acl-main.348} & 2020 & Modeling & \dsbox{seamecolor}  SEAME & zho-eng \\
\href{https://ieeexplore.ieee.org/document/9413806/}{10.1109/ICASSP39728.2021.9413806} & 2020 & Modeling & \dsbox{proprietarydatacolor} Proprietary Data & tha-eng \\
\href{https://ieeexplore.ieee.org/document/9413562/}{10.1109/ICASSP39728.2021.9413562} & 2020 & Modeling & \dsbox{seamecolor}  SEAME & zho-eng \\
\href{https://www.isca-archive.org/interspeech_2020/sharma20c_interspeech.html}{10.21437/interspeech.2020-2402} & 2020 & Data Augmentation & \dsbox{proprietarydatacolor} Proprietary Data & hin-eng \\
\href{https://www.isca-archive.org/interspeech_2020/li20l_interspeech.html}{10.21437/interspeech.2020-2177} & 2020 & Data Augmentation & \dsbox{seamecolor}  SEAME & zho-eng \\
\href{https://www.isca-archive.org/interspeech_2020/lu20f_interspeech.html}{10.21437/interspeech.2020-2485} & 2020 & Modeling & \dsbox{asru2019color} ASRU 2019 & zho-eng \\
\href{https://ieeexplore.ieee.org/document/9054138/}{10.1109/ICASSP40776.2020.9054138} & 2020 & Modeling & \dsbox{proprietarydatacolor} Proprietary Data & tam-eng \\
\href{https://ieeexplore.ieee.org/document/9383620/}{10.1109/SLT48900.2021.9383620} & 2020 & Data Augmentation & \dsbox{asru2019color} ASRU 2019 & zho-eng \\
\href{https://ieeexplore.ieee.org/document/9362075/}{10.1109/ISCSLP49672.2021.9362075} & 2020 & Modeling & \dsbox{seamecolor}  SEAME & zho-eng \\
\href{https://aclanthology.org/2020.lrec-1.523/}{aclanthology.org/2020.lrec-1.523/} & 2020 & Dataset & \dsbox{arzncolor} ArzEn & ara(egyptian)-eng \\
\href{https://www.isca-archive.org/interspeech_2020/zhou20b_interspeech.html}{10.21437/interspeech.2020-2488} & 2020 & Modeling & \dsbox{seamecolor}  SEAME & zho-eng \\
\href{https://ieeexplore.ieee.org/document/9058687/}{10.1109/ACCESS.2020.2986255} & 2020 & Modeling & \dsbox{iitghingcoscolor} IITG-HingCoS & hin-eng \\
\href{https://ieeexplore.ieee.org/document/9414428/}{10.1109/ICASSP39728.2021.9414428} & 2020 & Modeling & \dsbox{asru2019color} ASRU 2019 & zho-eng \\
\href{https://www.isca-archive.org/interspeech_2020/qiu20c_interspeech.html}{10.21437/interspeech.2020-1980} & 2020 & Modeling & \dsbox{seamecolor}  SEAME & zho-eng \\
\href{https://ieeexplore.ieee.org/document/9362080/}{10.1109/ISCSLP49672.2021.9362080} & 2021 & Modeling & \dsbox{seamecolor}  SEAME & zho-eng \\
\href{https://www.isca-archive.org/interspeech_2021/kumar21e_interspeech.html}{10.21437/interspeech.2021-978} & 2021 & Modeling & \dsbox{mucscolor} MUCS & hin-eng, ben-eng \\
\href{https://ieeexplore.ieee.org/document/9687941/}{10.1109/ASRU51503.2021.9687941} & 2021 & Dataset & \dsbox{germanspokenwikipediacorpuscolor}  German Spoken Wikipedia Corpus & deu-eng \\
\href{https://www.mdpi.com/2076-3417/11/19/9106}{10.3390/app11199106} & 2021 & Modeling & \dsbox{seamecolor}  SEAME & zho-eng \\
\href{https://ieeexplore.ieee.org/document/9747537/}{10.1109/icassp43922.2022.9747537} & 2021 & Modeling & \dsbox{asru2019color} ASRU 2019 & zho-eng \\
\href{https://asmp-eurasipjournals.springeropen.com/articles/10.1186/s13636-021-00222-7}{10.1186/s13636-021-00222-7} & 2021 & Other & \dsbox{proprietarydatacolor} Proprietary Data & zho-eng \\
\href{https://www.isca-archive.org/interspeech_2021/chowdhury21_interspeech.html}{10.21437/interspeech.2021-1809} & 2021 & Modeling & \dsbox{escwacolor} ESCWA, \dsbox{qasr.cscolor} QASR.CS & ara-eng, ara-fra \\
\href{https://www.isca-archive.org/interspeech_2021/sailor21_interspeech.html}{10.21437/interspeech.2021-1578} & 2021 & Modeling & \dsbox{mucscolor} MUCS & hin-eng, ben-eng \\
\href{https://linkinghub.elsevier.com/retrieve/pii/S0885230821000814}{10.1016/j.csl.2021.101278} & 2021 & Modeling & \dsbox{arzncolor} ArzEn & ara-eng \\
\href{https://www.isca-archive.org/interspeech_2021/ali21b_interspeech.html}{10.21437/interspeech.2021-2231} & 2021 & Modeling & \dsbox{escwacolor} ESCWA & ara-eng \\
\href{https://ieeexplore.ieee.org/document/9688174/}{10.1109/ASRU51503.2021.9688174} & 2021 & Modeling & \dsbox{seamecolor}  SEAME & zho-eng \\
\href{https://www.jstage.jst.go.jp/article/transinf/E104.D/10/E104.D_2021EDP7005/_article}{10.1587/transinf.2021edp7005} & 2021 & Modeling & \dsbox{proprietarydatacolor} Proprietary Data & jpn-eng \\
\href{https://www.isca-archive.org/interspeech_2021/diwan21_interspeech.html}{10.21437/Interspeech.2021-1339} & 2021 & Other & \dsbox{mucscolor} MUCS & hin-eng, ben-eng \\
\href{https://www.mdpi.com/2076-3417/11/6/2866}{10.3390/APP11062866} & 2021 & Modeling & \dsbox{proprietarydatacolor} Proprietary Data & kor-eng \\
\href{https://ieeexplore.ieee.org/document/9746830/}{10.1109/icassp43922.2022.9746830} & 2021 & Modeling & \dsbox{seamecolor}  SEAME & zho-eng \\
\href{https://www.jstage.jst.go.jp/article/transinf/E105.D/9/E105.D_2022EDL8036/_article}{10.1587/transinf.2022edl8036} & 2022 & Modeling & \dsbox{seamecolor}  SEAME & zho-eng \\
\href{https://www.isca-archive.org/interspeech_2022/shen22_interspeech.html}{10.21437/interspeech.2022-221} & 2022 & Modeling & \dsbox{proprietarydatacolor} Proprietary Data & zho-eng \\
\href{https://ieeexplore.ieee.org/document/10072473/}{10.1109/ICCT56141.2022.10072473} & 2022 & Modeling & \dsbox{proprietarydatacolor} Proprietary Data & yue-eng \\
\href{https://link.springer.com/10.1007/s11042-022-12136-3}{10.1007/s11042-022-12136-3} & 2022 & Modeling & \dsbox{proprietarydatacolor} Proprietary Data & zho-eng \\
\href{https://arxiv.org/abs/2206.13135}{10.21437/Interspeech.2022-877} & 2022 & Dataset & \dsbox{talcscorpuscolor}  TALCS corpus & zho-eng \\
\href{https://ieeexplore.ieee.org/document/10095878/}{10.1109/ICASSP49357.2023.10095878} & 2022 & Modeling & \dsbox{seamecolor}  SEAME & zho-eng \\
\href{https://ieeexplore.ieee.org/document/10037962/}{10.1109/ISCSLP57327.2022.10037962} & 2022 & Modeling & \dsbox{talcscorpuscolor}  TALCS corpus and \dsbox{magicdataramccolor} MagicData-RAMC & zho-eng \\
\href{https://www.tandfonline.com/doi/full/10.1080/09720529.2021.2014134}{10.1080/09720529.2021.2014134} & 2022 & Modeling & N.A & mar-eng \\
\href{https://ieeexplore.ieee.org/document/10037997/}{10.1109/ISCSLP57327.2022.10037997} & 2022 & Modeling & \dsbox{proprietarydatacolor} Proprietary Data & zho-en \\
\href{https://link.springer.com/10.1007/s10579-022-09592-6}{10.1007/s10579-022-09592-6} & 2022 & Dataset & \dsbox{twosepedienglishcsspeechcorpus} Two Sepedi Corpus & nso-eng \\
\href{https://ieeexplore.ieee.org/document/10038051/}{10.1109/ISCSLP57327.2022.10038051} & 2022 & Other & \dsbox{talcscorpuscolor}  TALCS corpus and \dsbox{magicdataramccolor} MagicData-RAMC & zho-eng \\
\href{https://arxiv.org/abs/2206.02093}{10.21437/Interspeech.2022-923} & 2022 & Modeling & \dsbox{asru2019color} ASRU 2019 & zho-eng \\
\href{https://ieeexplore.ieee.org/document/10097151/}{10.1109/ICASSP49357.2023.10097151} & 2022 & Modeling & \dsbox{seamecolor}  SEAME & zho-eng \\
\href{https://www.isca-archive.org/interspeech_2022/zhang22x_interspeech.html}{10.21437/interspeech.2022-10286} & 2022 & Modeling & \dsbox{asru2021color} ASRU 2021 & zho-eng \\
\href{https://www.isca-archive.org/interspeech_2022/antony22_interspeech.html}{10.21437/interspeech.2022-10763} & 2022 & Modeling & \dsbox{mucscolor} MUCS & hin-eng \\
\href{https://arxiv.org/abs/2206.14580}{10.21437/Interspeech.2022-11426} & 2022 & Modeling & \dsbox{aicubescompetitiondatasetcolor} 
 AIcubes competition dataset & zho-eng \\
\href{https://ieeexplore.ieee.org/document/10038194/}{10.1109/ISCSLP57327.2022.10038194} & 2022 & Modeling & \dsbox{iscslp2022magichubcodeswitchingasrchallengedatasetcolor} ISCSLP 2022 Code-Switching Challenge & zho-eng \\
\href{https://www.isca-archive.org/interspeech_2022/ye22_interspeech.html}{10.21437/interspeech.2022-719} & 2022 & Modeling & \dsbox{proprietarydatacolor} Proprietary Data & zho-eng \\
\href{https://ieeexplore.ieee.org/document/10023181/}{10.1109/SLT54892.2023.10023181} & 2022 & Evaluation Metric & \dsbox{haccolor} HAC & ara-eng \\
\href{https://www.isca-archive.org/interspeech_2022/mussakhojayeva22_interspeech.html}{10.21437/interspeech.2022-421} & 2022 & Dataset & KSC2 & kaz-eng \\
\href{https://arxiv.org/abs/2307.05956}{10.21437/Interspeech.2023-2292} & 2023 & Modeling & \dsbox{asru2019color} ASRU 2019 & zho-eng \\
\href{https://arxiv.org/abs/2305.11095}{10.21437/Interspeech.2023-2032} & 2023 & Other & \dsbox{ascendcolor} ASCEND, \dsbox{seamecolor}  SEAME & zho-eng \\
\href{https://aclanthology.org/2023.findings-emnlp.543}{10.18653/v1/2023.findings-emnlp.543} & 2023 & Data Augmentation & \dsbox{seamecolor}  SEAME, \dsbox{ascendcolor} ASCEND & zho-eng \\
\href{https://ieeexplore.ieee.org/document/10317410/}{10.1109/APSIPAASC58517.2023.10317410} & 2023 & Modeling & \dsbox{ascendcolor} ASCEND, \dsbox{ntutab01color} NTUT-AB01 & zho-eng \\
\href{https://ieeexplore.ieee.org/document/10389644/}{10.1109/ASRU57964.2023.10389644} & 2023 & Data Augmentation & \dsbox{seamecolor}  SEAME & zho-eng \\
\href{https://www.isca-archive.org/interspeech_2023/chi23_interspeech.html}{10.21437/interspeech.2023-1050} & 2023 & Data Augmentation & \dsbox{seamecolor}  SEAME & zho-eng \\
\href{https://ieeexplore.ieee.org/document/10448335/}{10.1109/ICASSP48485.2024.10448335} & 2023 & Modeling & \dsbox{asru2019color} ASRU 2019 & zho-eng \\
\href{https://ieeexplore.ieee.org/document/10346710/}{10.1109/ICEEI59426.2023.10346710} & 2023 & Modeling & \dsbox{hartantocscolor} Hartanto CS & ind-eng \\
\href{https://ieeexplore.ieee.org/document/10096429/}{10.1109/ICASSP49357.2023.10096429} & 2023 & Modeling & \dsbox{proprietarydatacolor} Proprietary Data & hin-eng \\
\href{https://ieeexplore.ieee.org/document/10096317/}{10.1109/ICASSP49357.2023.10096317} & 2023 & Modeling & \dsbox{proprietarydatacolor} Proprietary Data & zho-eng \\
\href{https://www.mdpi.com/2076-3417/13/14/8492}{10.3390/app13148492} & 2023 & Dataset & Bilingual Basque-Spanish Dataset & eus-spa \\
\href{https://ieeexplore.ieee.org/document/10446258/}{10.1109/ICASSP48485.2024.10446258} & 2023 & Modeling & \dsbox{seamecolor}  SEAME & zho-eng \\
\href{https://aclanthology.org/2023.calcs-1.7}{10.18653/v1/2023.calcs-1.7} & 2023 & Modeling & \dsbox{mucscolor} MUCS & hin-eng, spa-eng \\
\href{https://www.isca-archive.org/interspeech_2023/liang23b_interspeech.html}{10.21437/interspeech.2023-923} & 2023 & Data Augmentation & \dsbox{asru2019color} ASRU 2019 & zho-eng \\
\href{https://arxiv.org/abs/2311.15077}{aclanthology.org/2023.calcs-1.8/} & 2023 & Modeling & \dsbox{southafricansoapoperascolor} South African Soap Operas & eng-(zul, xho, sot, tsn) \\
\href{https://ieeexplore.ieee.org/document/10274649/}{10.1109/LSP.2023.3307350} & 2023 & Modeling & \dsbox{seamecolor}  SEAME & zho-eng \\
\href{https://ieeexplore.ieee.org/document/10445734/}{10.1109/ICASSP48485.2024.10445734} & 2023 & Modeling, Dataset & \dsbox{tunswitchcolor} TunSwitch & ara(tunisian)-fra-eng \\
\href{https://www.isca-archive.org/interspeech_2023/fan23_interspeech.html}{10.21437/interspeech.2023-262} & 2023 & Modeling & \dsbox{seamecolor}  SEAME & zho-eng \\
\href{https://aclanthology.org/2023.calcs-1.4}{10.18653/v1/2023.calcs-1.4} & 2023 & Modeling & \dsbox{seamecolor}  SEAME & zho-eng \\
\href{https://ieeexplore.ieee.org/document/10389662/}{10.1109/ASRU57964.2023.10389662} & 2023 & Modeling & \dsbox{asru2019color} ASRU 2019 & zho-eng \\
\href{https://ieeexplore.ieee.org/document/10022475/}{10.1109/SLT54892.2023.10022475} & 2023 & Modeling & \dsbox{proprietarydatacolor} Proprietary Data & hin-eng \\
\href{https://www.isca-archive.org/interspeech_2023/tan23c_interspeech.html}{10.21437/interspeech.2023-1465} & 2023 & Modeling & \dsbox{asru2019color} ASRU 2019 & zho-eng \\
\href{https://ieeexplore.ieee.org/document/10389798/}{10.1109/ASRU57964.2023.10389798} & 2023 & Modeling & \dsbox{asru2019color} ASRU 2019 & zho-eng \\
\href{https://ieeexplore.ieee.org/document/10446857/}{10.1109/ICASSP48485.2024.10446857} & 2023 & Data Augmentation & \dsbox{seamecolor}  SEAME, \dsbox{escwacolor} ESCWA & zho-eng, ara-eng \\
\href{https://ieeexplore.ieee.org/document/10170166/}{10.1109/ICEIB57887.2023.10170166} & 2023 & Modeling & \dsbox{talcscorpuscolor}  TALCS corpus & zho-eng \\
\href{https://linkinghub.elsevier.com/retrieve/pii/S0003682X24002706}{10.1016/j.apacoust.2024.110119} & 2024 & Dataset & \dsbox{vitbhebiccolor} VITB-HEBiC & hin-eng \\
\href{https://ieeexplore.ieee.org/document/10446652/}{10.1109/ICASSP48485.2024.10446652} & 2024 & Modeling & \dsbox{seamecolor}  SEAME & zho-eng \\
\href{https://linkinghub.elsevier.com/retrieve/pii/S088523082400010X}{10.1016/j.csl.2024.101627} & 2024 & Dataset & MECOS & mni-eng \\
\href{https://aclanthology.org/2024.lrec-main.174/}{aclanthology.org/2024.lrec-main.174/} & 2024 & Other & \dsbox{southafricansoapoperascolor} South African Soap Operas & eng - (zul, xho, sot, tsn, nso) \\
\href{https://arxiv.org/abs/2405.02578}{aclanthology.org/2024.sigul-1.26/} & 2024 & Dataset & \dsbox{mixatcolor} Mixat & ara(emirati)-eng \\
\href{https://aclanthology.org/2024.lrec-main.400/}{aclanthology.org/2024.lrec-main.400/} & 2024 & Dataset & \dsbox{decmcolor} DECM & deu-eng \\
\href{https://arxiv.org/abs/2404.15501}{aclanthology.org/2024.lrec-main.852/} & 2024 & Dataset & Kilkarn & que-spa \\
\href{https://aclanthology.org/2024.sigul-1.18.pdf}{aclanthology.org/2024.sigul-1.18.pdf} & 2024 & Modeling & \dsbox{hartantocscolor} Hartanto CS & ind-eng \\
\href{https://arxiv.org/abs/2403.18182}{aclanthology.org/2024.lrec-main.1546/} & 2024 & Dataset & \dsbox{zaebucspokencolor} ZAEBUC-Spoken & ara-eng, ara-ara \\
\href{https://linkinghub.elsevier.com/retrieve/pii/S0003682X24000343}{10.1016/j.apacoust.2024.109883} & 2024 & Modeling & \dsbox{talcscorpuscolor}  TALCS corpus & zho-eng \\
\href{https://dl.acm.org/doi/10.1145/3640794.3665579}{10.1145/3640794.3665579} & 2024 & Modeling & \dsbox{bangormiamicolor} Bangor Miami & spa-eng \\
\href{https://aclanthology.org/2024.lrec-main.262/}{aclanthology.org/2024.lrec-main.262/} & 2024 & Modeling & \dsbox{proprietarydatacolor} Proprietary Data & zho-eng \\
\href{https://ieeexplore.ieee.org/document/10832233/}{10.1109/SLT61566.2024.10832233} & 2024 & Other & \dsbox{seamecolor}  SEAME & zho-eng \\
\href{https://ieeexplore.ieee.org/document/10832265/}{10.1109/SLT61566.2024.10832265} & 2024 & Modeling & \dsbox{asru2019color} ASRU 2019 & zho-eng \\
\href{https://arxiv.org/abs/2412.16507}{10.1109/ICASSP49660.2025.10889634} & 2024 & Modeling & \dsbox{seamecolor}  SEAME & zho-eng \\
\href{https://ieeexplore.ieee.org/document/10849279/}{10.1109/apsipaasc63619.2025.10849279} & 2024 & Modeling & \dsbox{talcscorpuscolor}  TALCS corpus & zho-eng \\
\href{https://ieeexplore.ieee.org/document/10750818/}{10.1109/ACCESS.2024.3496617} & 2024 & Modeling & Roman Urdu code-mixed dataset & urd-eng \\
\href{https://aclanthology.org/2024.findings-emnlp.356}{10.18653/v1/2024.findings-emnlp.356} & 2024 & Evaluation Metric & \dsbox{mixatcolor} Mixat & ara(emirati)-eng \\
\href{https://www.isca-archive.org/interspeech_2024/ye24_interspeech/}{10.21437/Interspeech.2024-259} & 2024 & Modeling & \dsbox{asru2019color} ASRU 2019 & zho-eng \\
\href{https://aclanthology.org/2024.emnlp-main.1211/}{10.18653/v1/2024.emnlp-main.1211} & 2024 & Dataset & \dsbox{casablancacolor} Casablanca & ara-fra, ara-eng \\
\href{https://ieeexplore.ieee.org/document/10832290/}{10.1109/SLT61566.2024.10832290} & 2024 & Modeling & \dsbox{ascendcolor} ASCEND & zho-eng \\
\href{https://aclanthology.org/2024.lrec-main.308/}{aclanthology.org/2024.lrec-main.308/} & 2024 & Data Augmentation & \dsbox{proprietarydatacolor} Proprietary Data & lav-eng \\
\href{https://ieeexplore.ieee.org/document/10832173/}{10.1109/SLT61566.2024.10832173} & 2024 & Other & \dsbox{asru2019color} ASRU 2019 & zho-eng \\
\href{https://www.isca-archive.org/interspeech_2024/hussein24_interspeech.html}{10.21437/interspeech.2024-1418} & 2024 & Modeling & \dsbox{seamecolor}  SEAME & zho-eng \\
\href{https://ieeexplore.ieee.org/document/10796602/}{10.1109/COMPAS60761.2024.10796602} & 2024 & Modeling & \dsbox{proprietarydatacolor} Proprietary Data & ben-eng \\
\href{https://arxiv.org/abs/2406.18120}{10.1016/j.procs.2024.10.184} & 2024 & Modeling & \dsbox{arzncolor} ArzEn & ara(egyptian)-eng \\
\href{https://ieeexplore.ieee.org/document/10832326/}{10.1109/SLT61566.2024.10832326} & 2024 & Modeling & \dsbox{seamecolor}  SEAME & zho-eng \\
\href{https://www.isca-archive.org/syndata4genai_2024/alharbi24_syndata4genai.html}{10.21437/syndata4genai.2024-6} & 2024 & Dataset & \dsbox{saudilangcorpus(scc)(evalonly)color} Saudi Lang Corpus (SCC) (Eval only)  & ara(saudi)-en \\
\href{https://ieeexplore.ieee.org/document/10634607/}{10.1109/ICCE62051.2024.10634607} & 2024 & Modeling & \dsbox{seamecolor}  SEAME & zho-eng \\
\href{https://ieeexplore.ieee.org/document/10890805}{10.1109/ICASSP49660.2025.10890805} & 2024 & Modeling & \dsbox{asru2019color} ASRU 2019 & zho-eng \\
\href{https://arxiv.org/abs/2406.03814}{10.1109/ICASSP49660.2025.10890024} & 2024 & Modeling & \dsbox{ascendcolor} ASCEND & zho-eng \\
\href{https://ieeexplore.ieee.org/document/10627333/}{10.1109/ICASSPW62465.2024.10627333} & 2024 & Modeling & \dsbox{kecscolor} KECS & kor-eng  \\
\bottomrule
\caption{Full list of surveyed papers. \textbf{Year} corresponds the earliest available version, which can be a preprint. }
\end{longtable}

\end{document}